\documentclass[10pt,twocolumn,letterpaper]{article}

\usepackage{times}
\usepackage{latexsym}

\usepackage{algorithm}
\usepackage{algpseudocode}
\usepackage{amsmath}
\usepackage{amssymb}
\usepackage{booktabs}
\usepackage{multirow}
\usepackage{graphicx}
\usepackage{adjustbox}
\usepackage[table]{xcolor}
\definecolor{ourblue}{RGB}{235,245,255}
\usepackage{placeins}
\usepackage{needspace}
\usepackage[most]{tcolorbox}
\usepackage{makecell}
\usepackage{booktabs}
\usepackage{tabularx}
\usepackage{array}
\usepackage{pifont}

\newcommand{\cmark}{\ding{51}}
\newcommand{\xmark}{\ding{55}}

\usepackage{wacv}              

\definecolor{wacvblue}{rgb}{0.21,0.49,0.74}
\usepackage[pagebackref,breaklinks,colorlinks,allcolors=wacvblue]{hyperref}

\def\wacvPaperID{526} 
\def\confName{WACV}
\def\confYear{2027}

\title{CatchUp: Depth-Aware Re-Admission of Visual Tokens \\ for Efficient Large Vision-Language Models}

\author{
    Eunju Park \quad \quad \quad 
    Uiwon Hwang\thanks{Corresponding author.} \\
    Division of Artificial Intelligence \& Software, Ewha Womans University \\
    {\tt\small \{pej0918, uiwon.hwang\}@ewha.ac.kr}
}

\begin{document}
\maketitle

\begin{abstract}
Large Vision-Language Models (LVLMs) repeatedly process hundreds of visual tokens across decoder layers, incurring substantial computation and memory costs.
Existing visual-token reduction methods often irreversibly discard tokens despite depth-varying query relevance.
Although recoverable routing allows excluded tokens to return, re-admitted tokens retain stale hidden states after bypassing intermediate decoder transformations.
Thus, token recoverability does not guarantee representation readiness at the current decoder depth.
Under a fixed routing trajectory at a 64-token budget, replacing stale re-admitted states with current-depth oracle representations improves average RefCOCO Acc@0.5 from 18.96 to 23.18, revealing the impact of representation staleness.
We propose \textbf{CatchUp}, a training-free framework combining reversible visual-token routing with \textbf{Local Transition Correction (LTC)}.
CatchUp repeatedly re-evaluates query-conditioned relevance over the complete visual-token set, preserving deferred tokens for later re-admission.
LTC approximates their skipped representation evolution by transferring reference-to-current hidden-state transitions from semantically and spatially related active anchors.
This correction improves alignment with current-depth oracle representations without recomputing the skipped decoder layers.
Across three referring-expression grounding datasets and six multimodal benchmarks, CatchUp achieves the best grounding performance across all eight splits and the highest average multimodal retention among evaluated efficient methods at 64-, 128-, and 192-token budgets.
At a 64-token budget, CatchUp processes 88.9\% fewer active visual tokens per post-routing decoder interval, retains 92.17\% of dense-model multimodal performance, and reduces end-to-end latency by 8.0\%.
Code is available at \href{https://github.com/pej0918/CatchUp}{https://github.com/pej0918/CatchUp}.
\end{abstract}
    
\section{Introduction}
\label{sec:introduction}

\begin{table*}[t]
    \centering
    \footnotesize
    \setlength{\tabcolsep}{5.0pt}
    \renewcommand{\arraystretch}{1.10}
    \begin{tabularx}{\textwidth}{
        @{}
        >{\raggedright\arraybackslash}p{2.15cm}
        >{\raggedright\arraybackslash}X
        >{\centering\arraybackslash}p{1.60cm}
        >{\centering\arraybackslash}p{2.00cm}
        >{\centering\arraybackslash}p{1.75cm}
        @{}
    }
        \toprule
        \textbf{Method}
        & \textbf{Routing and token handling}
        & \shortstack{\textbf{Repeated}\\\textbf{selection}}
        & \shortstack{\textbf{Original-token}\\\textbf{recoverability}}
        & \shortstack{\textbf{Re-admission}\\\textbf{correction}} \\
        \midrule
        FastV~\cite{chen2024fastv}
        & One-shot permanent pruning
        & \xmark & \xmark & \xmark \\
        SparseVLM~\cite{zhang2025sparsevlm}
        & Adaptive pruning and recycling
        & \cmark & \xmark & \xmark \\
        PyramidDrop~\cite{xing2025pyramiddrop}
        & Progressive stage-wise pruning
        & \cmark & \xmark & \xmark \\
        N\"uwa~\cite{huang2026nuwa}
        & Spatial aggregation and pruning
        & \xmark & \xmark & \xmark \\
        VScan~\cite{zhang2026vscan}
        & Encoder-side merging and intermediate-layer pruning
        & \xmark & \xmark & \xmark \\
        ALVTS~\cite{chen2026alvts}
        & Per-layer bypass and reintegration
        & \cmark & \cmark & \xmark \\
        ReRoute~\cite{yang2026reroute}
        & Stage-wise deferral and reintegration
        & \cmark & \cmark & \xmark \\
        \midrule
        \textbf{CatchUp}
        & Reversible routing with depth-aware re-admission
        & \cmark & \cmark & \cmark \\
        \bottomrule
    \end{tabularx}
    \caption{
        Structural comparison of representative visual-token reduction methods.
        Repeated selection denotes reassessment of visual-token participation at multiple \emph{decoder} depths.
        Original-token recoverability indicates whether excluded tokens remain independently eligible for later computation.
        Re-admission correction indicates explicit compensation for decoder transformations skipped by returning tokens.
    }
    \label{tab:method_comparison}
\end{table*}

Large Vision-Language Models (LVLMs) process hundreds or thousands of visual tokens across decoder layers, incurring substantial computation and memory costs.
Visual-token reduction alleviates this overhead by compressing visual representations before decoding or reducing tokens within the decoder~\cite{yang2025visionzip,chen2024fastv,zhang2025sparsevlm,xing2025pyramiddrop,huang2026nuwa}.

Most decoder-side methods prune or merge visual tokens based on relevance, redundancy, or spatial structure~\cite{chen2024fastv,zhang2025sparsevlm,xing2025pyramiddrop,huang2026nuwa}.
However, visual tokens receiving high query-conditioned attention can change across decoder depth, while irreversibly removed tokens cannot return when ranked highly at later layers.
Recoverable routing addresses this limitation by preserving skipped tokens for subsequent reconsideration~\cite{chen2026alvts,yang2026reroute}.

\emph{Token recoverability, however, does not guarantee representation readiness.}
Deferred tokens bypass intermediate decoder transformations and retain earlier hidden states, potentially becoming misaligned with the current decoder depth upon re-admission.
Existing recoverable methods enable tokens to return but do not explicitly compensate for their skipped representation evolution.
Thus, recovering visual evidence and restoring its depth-aligned representation are distinct challenges.

We empirically examine both challenges.
Under full-token inference, the visual tokens favored by query-conditioned attention change substantially across decoder depth, motivating repeated evidence reconsideration.
Moreover, replacing stale re-admitted states with current-depth oracle representations under a fixed routing trajectory improves average RefCOCO accuracy from 18.96 to 23.18 at a 64-token budget.
This controlled intervention shows that re-admitted representation quality affects grounding even when token-selection decisions are held fixed.

We propose \textbf{CatchUp}, a training-free framework combining reversible visual-token routing with \textbf{Local Transition Correction (LTC)}.
CatchUp repeatedly evaluates query-conditioned relevance over the complete visual-token set, processes selected tokens between routing layers, and preserves deferred tokens for later re-admission.
Upon re-admission, LTC estimates the skipped representation evolution using reference-to-current hidden-state transitions from semantically and spatially related active anchors.
The estimated transition is applied to the token's own cached state, improving depth alignment without processing deferred tokens through the skipped decoder layers.

Across three referring-expression grounding datasets and six multimodal benchmarks, CatchUp achieves the best grounding performance among evaluated efficient methods
at 64-, 128-, and 192-token budgets.
LTC improves current-depth oracle alignment, increasing cosine similarity from 0.6299 to 0.7373.
At $K=64$, CatchUp processes 88.9\% fewer active visual tokens per post-routing decoder interval, retains 92.17\% of dense-model multimodal performance,
and reduces end-to-end latency by 8.0\%.

Our main contributions are:
\begin{itemize}
    \item We identify two limitations of visual-token reduction: irreversible loss of late-emerging evidence and representation staleness after re-admission, supported by layer-wise relevance analysis and a controlled oracle intervention.

    \item We introduce \textbf{CatchUp}, combining reversible routing with \textbf{LTC}, which approximates skipped hidden-state transitions through semantic--spatial anchor transfer without full-token computation.

    \item We demonstrate that LTC improves current-depth representation alignment across re-admission gaps and consistently enhances grounding performance, while maintaining favorable multimodal retention and inference efficiency.
\end{itemize}

\section{Related Work}
\label{sec:related_work}

\subsection{Irreversible Visual Token Reduction}

Visual-token reduction methods operate either before decoding
or within the language decoder.
Pre-decoder methods select or merge redundant vision-encoder
outputs~\cite{yang2025visionzip}, while decoder-side methods
exploit attention or query relevance~\cite{chen2024fastv,zhang2025sparsevlm},
adaptive token budgets~\cite{ye2025fitprune,ye2025atpllava,yang2025topv},
progressive reduction~\cite{xing2025pyramiddrop},
diversity and duplication~\cite{alvar2025divprune,wen2025dart},
or spatial and grounding constraints~\cite{endo2025feather,chien2025gap,huang2026nuwa}.
VScan~\cite{zhang2026vscan} combines encoder-side merging
with intermediate-layer decoder pruning.
Despite their different reduction strategies, these methods
do not preserve excluded original visual tokens for
independent re-admission at later decoder depths.
Consequently, visual evidence discarded early cannot be
reconsidered when its query relevance increases at later layers.

\subsection{Recoverable Visual Token Routing}

Recoverable routing addresses irreversible evidence loss
by allowing excluded visual tokens to return to decoder computation.
ALVTS~\cite{chen2026alvts} performs adaptive per-layer
selection, bypassing computation for unselected tokens
before reintegrating them for subsequent selection.
ReRoute~\cite{yang2026reroute} instead employs stage-wise
deferral and reconsideration, preserving deferred tokens
as candidates for later routing decisions.
Both approaches are related to token-wise conditional computation
in Mixture-of-Depths~\cite{raposo2024mixtureofdepths},
but specifically target visual tokens in pretrained LVLM decoders.

Although ALVTS and ReRoute differ in routing granularity,
both enable previously excluded visual tokens to resume computation.
However, their emphasis is on token participation and
recoverability rather than representation alignment upon re-admission.
Returning tokens retain earlier-depth hidden states because
they bypass intermediate decoder transformations.
Neither method explicitly compensates for this skipped
representation evolution, leaving a potential mismatch
between re-admitted and continuously active tokens.

\paragraph{Distinction from prior work.}
As summarized in \Cref{tab:method_comparison},
CatchUp builds on reversible routing but addresses a distinct
limitation: \emph{token recoverability does not guarantee
representation readiness}.
Rather than merely restoring deferred tokens to computation,
CatchUp introduces Local Transition Correction (LTC),
which estimates their missed representation evolution
from reference-to-current hidden-state transitions of
semantically and spatially related active anchors.
By applying the estimated transition to each returning
token's own cached state, CatchUp explicitly addresses
the representation mismatch caused by skipped decoder layers.
Thus, CatchUp extends recoverable routing from deciding
\emph{which} tokens can return to correcting \emph{how}
they resume computation at the current decoder depth.

\section{Motivating Observations}
\label{sec:observations}

To identify the requirements for effective recoverable
visual-token routing, we investigate two questions:
(i) Can early-excluded visual tokens become highly
ranked by the same query-conditioned attention criterion
at later decoder depths?
(ii) Does representation staleness affect downstream
grounding when re-admission decisions are held fixed?


\subsection{Observation I: Query-Relevant Visual-Token Sets Change Across Decoder Depth}
\label{sec:observation_evidence}

\begin{figure}[t]
    \centering
    \includegraphics[width=0.98\columnwidth]
    {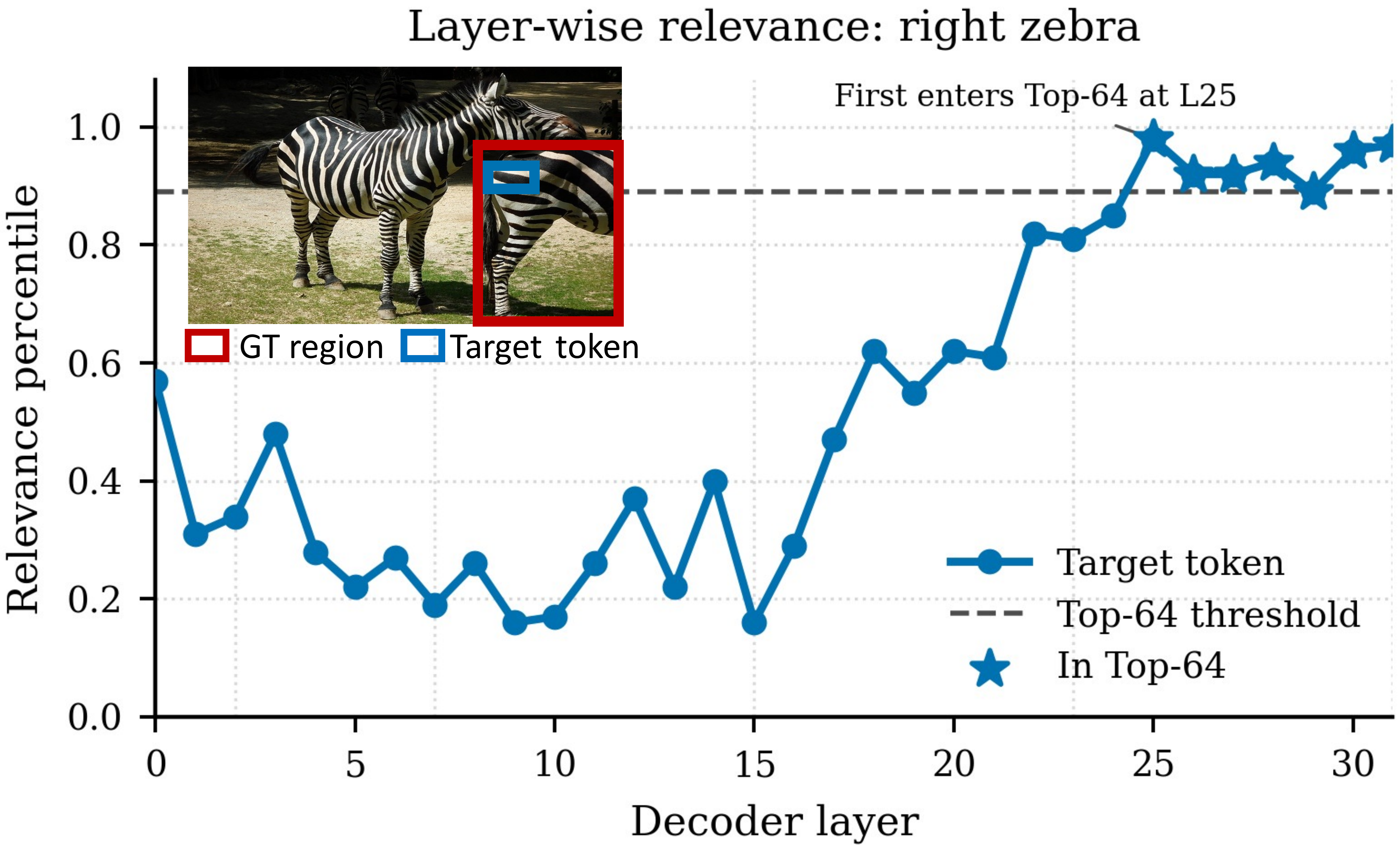}
    \caption{
    Late-emerging visual evidence for \emph{right zebra}.
    The target region (red) contains a token (blue) that
    first enters the top-64 relevance set at layer 25.
    The curve shows its relevance percentile; the dashed
    line marks the top-64 threshold, and stars denote
    top-64 membership.
    }
    \label{fig:zebra_late_emerging}
\end{figure}

\paragraph{Analysis protocol.}
We analyze LLaVA-1.5-7B~\cite{liu2024llava15} on
500 RefCOCO samples~\cite{yu2016modeling} under
full-token inference with 576 visual tokens.
At each decoder layer, query-conditioned relevance
is measured using post-softmax attention from the
last text token to each visual token, averaged
over attention heads.
For sample $n$, let $\mathcal E_n^\ell$ denote the
top-$K$ visual-token set at layer $\ell$, with $K=64$.
We use layer 2, the first routing layer in CatchUp,
as the reference depth.

\Cref{fig:zebra_late_emerging} illustrates a target-region
token that becomes highly ranked only at a late
decoder depth.
This example suggests that early selection can miss
visual evidence favored by the same relevance criterion
at later layers.

\paragraph{Dataset-level evidence dynamics.}
To examine whether such changes are systematic,
we measure the variation of top-$K$ token sets
across decoder depth.
We define \emph{relevance volatility} as the fraction
of top-$K$ tokens that change between adjacent layers:
\begin{equation}
\operatorname{Vol}_n(\ell)
=
1-
\frac{
    \left|
        \mathcal{E}_n^\ell
        \cap
        \mathcal{E}_n^{\ell-1}
    \right|
}{K}.
\label{eq:volatility}
\end{equation}

Volatility captures local changes in relevance,
but not how much later-ranked evidence would be
excluded by a single early selection.
We therefore define the \emph{late-emerging
evidence ratio} (LER) relative to
$\ell_{\mathrm{ref}}=2$:
\begin{equation}
\operatorname{LER}_n(\ell)
=
\frac{
    \left|
        \mathcal{E}_n^\ell
        \setminus
        \mathcal{E}_n^{\ell_{\mathrm{ref}}}
    \right|
}{K}.
\label{eq:ler}
\end{equation}

LER measures the fraction of the current top-$K$
set absent from the layer-2 selection.
Under irreversible layer-2 pruning, these tokens
would no longer be available for reconsideration.

\begin{figure}[t]
    \centering
    \includegraphics[width=0.86\columnwidth]
    {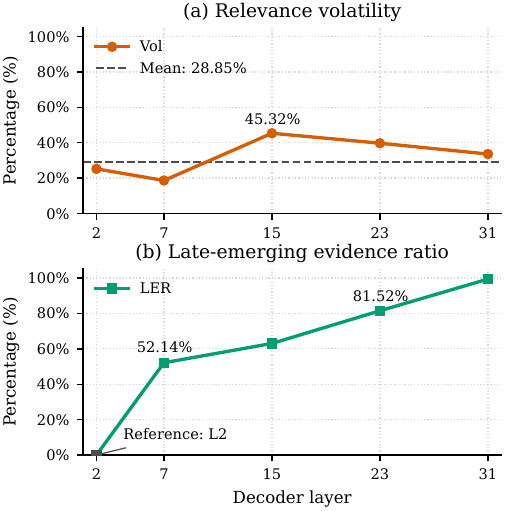}
    \caption{
    Evidence dynamics over 500 RefCOCO samples.
    (a) Mean relevance volatility is 28.85\% across
    adjacent layers (dashed line).
    (b) LER relative to layer 2 reaches 99.42\%
    at layer 31.
    }
    \label{fig:evidence_dynamics}
\end{figure}

As shown in \Cref{fig:evidence_dynamics}, query-relevant
token sets change substantially across depth,
both between neighboring layers and relative
to an early selection.
LER already exceeds 50\% at layer 7,
indicating that much of the later-ranked evidence
would be unavailable under irreversible early pruning.

These results do not establish that later-layer
relevance estimates are inherently more accurate.
Rather, they show that a single early selection
cannot preserve every token that may become
highly ranked at later depths.
This motivates preserving excluded visual tokens
for subsequent reconsideration.
The remaining question is whether their
representations are ready for computation
when they return.


\subsection{Observation II: Re-Admitted Tokens Carry Stale Representations}
\label{sec:observation_staleness}

Deferred tokens bypass intermediate self-attention
and feed-forward transformations, retaining
earlier-depth representations while continuously
active tokens undergo further updates.
This creates a potential representation mismatch
when deferred tokens are re-admitted.

We examine whether stale representations remain
informative for routing decisions and whether
they are suitable for downstream computation
after re-admission.

\paragraph{Stale representations retain useful routing information.}

\begin{table}[t]
\centering
\scriptsize
\renewcommand{\arraystretch}{1.05}
\begin{adjustbox}{max width=\columnwidth}
\begin{tabular}{c l cc}
\toprule
\textbf{Budget}
& \textbf{Routing}
& \textbf{Top-$K$ Overlap (\%) $\uparrow$}
& \textbf{Re-adm. Recall@K (\%) $\uparrow$} \\
\midrule
64
& Random
& $10.50{\pm}0.08$
& $10.59{\pm}0.20$ \\
& Stale
& \textbf{59.28}
& \textbf{38.67} \\
\midrule
128
& Random
& $19.82{\pm}0.07$
& $19.70{\pm}0.09$ \\
& Stale
& \textbf{62.04}
& \textbf{43.56} \\
\midrule
192
& Random
& $33.72{\pm}0.08$
& $33.37{\pm}0.18$ \\
& Stale
& \textbf{71.31}
& \textbf{46.33} \\
\bottomrule
\end{tabular}
\end{adjustbox}
\caption{
Stale-state routing agreement with current-depth
oracle routing.
Across three budgets, stale routing achieves
59--71\% top-$K$ overlap and 39--46\%
re-admission Recall@K, substantially exceeding
uniform random selection.
Random results are reported as mean $\pm$
standard deviation over five seeds.
}
\label{tab:stale_routing_agreement}
\end{table}

We compare stale and current-depth oracle routing
using the same current-layer text query and key
projection, differing only in the visual-token state:
the cached stale representation or the corresponding
current-depth representation from a separate
full-token trajectory.
Oracle scores are used only for diagnosis
and do not alter the stale-routing trajectory.

Let $\mathcal V_{\mathrm{stale}}^s$ and
$\mathcal V_{\mathrm{oracle}}^s$ denote the resulting
top-$K$ selections.
We measure Top-$K$ overlap as
$|\mathcal V_{\mathrm{stale}}^s
\cap\mathcal V_{\mathrm{oracle}}^s|/K$,
and re-admission Recall@K as the fraction of
oracle-selected deferred tokens also selected
by stale routing.

\Cref{tab:stale_routing_agreement} shows that
stale-state routing achieves substantially greater
agreement with current-depth oracle selection
than random routing across all budgets.
Nevertheless, the imperfect agreement indicates
that cached states provide useful but incomplete
signals for reconsideration.
More importantly, their utility for routing
does not guarantee representation readiness
for subsequent decoder computation.

\paragraph{Oracle intervention.}
To isolate the downstream effect of representation
staleness, we hold the routing trajectory and
visual-token budget fixed, modifying only the
hidden states of newly re-admitted tokens.
All other token states remain unchanged at
each intervention point.

For each newly re-admitted token $j$ at routing
layer $\ell_s$, we compare its stored
\emph{stale} state with its \emph{current-depth
oracle} state $h_{j,\mathrm{full}}[\ell_s]$
obtained from a separate full-token
reference trajectory.

\begin{figure}[t]
    \centering
    \includegraphics[width=0.90\columnwidth]
    {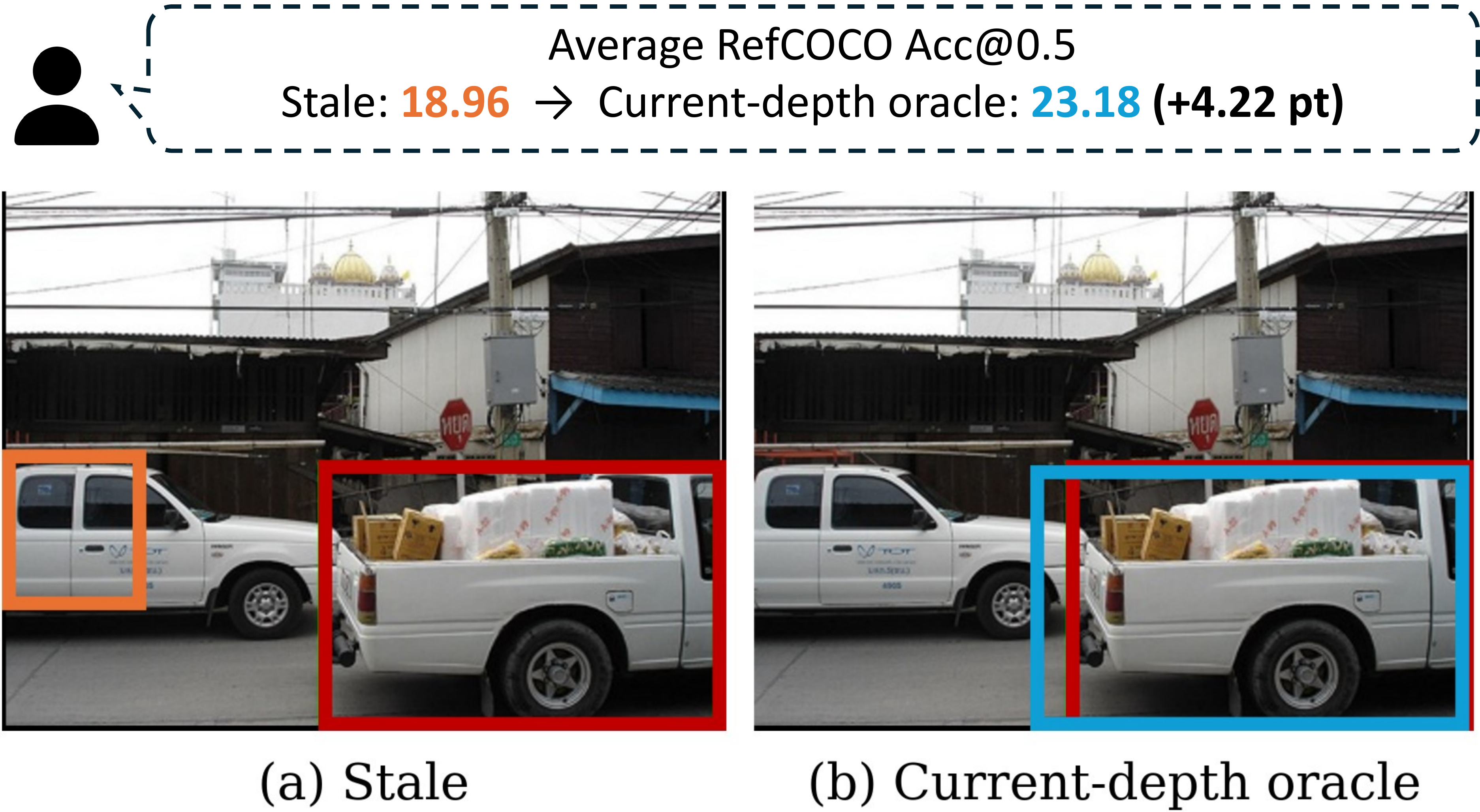}
    \caption{
    Fixed-routing oracle intervention on
    re-admitted token states.
    Replacing stale states with current-depth oracle
    states improves average RefCOCO Acc@0.5
    from 18.96 to 23.18.
    For \emph{right truck}, the red box marks
    the ground truth; orange (a) and blue (b)
    indicate stale and oracle predictions.
    }
    \label{fig:oracle_grounding_example}
\end{figure}

\Cref{fig:oracle_grounding_example} shows that
replacing stale states with current-depth oracle
representations improves grounding without
changing token-selection decisions.
The qualitative example likewise shows that
the oracle condition localizes the target
more accurately than the stale condition.

The improvement under fixed routing decisions provides direct evidence that re-admitted representation quality affects grounding beyond token-selection decisions.
Together, these observations motivate two complementary requirements: \emph{evidence recoverability} across decoder depths and \emph{representation readiness}
upon re-admission.
Since obtaining current-depth oracle states requires
full-token computation, efficient correction must
approximate the representation evolution missed
during deferral.

\section{Method}
\label{sec:method}

\begin{figure*}[t]
    \centering
    \includegraphics[width=0.98\textwidth]
    {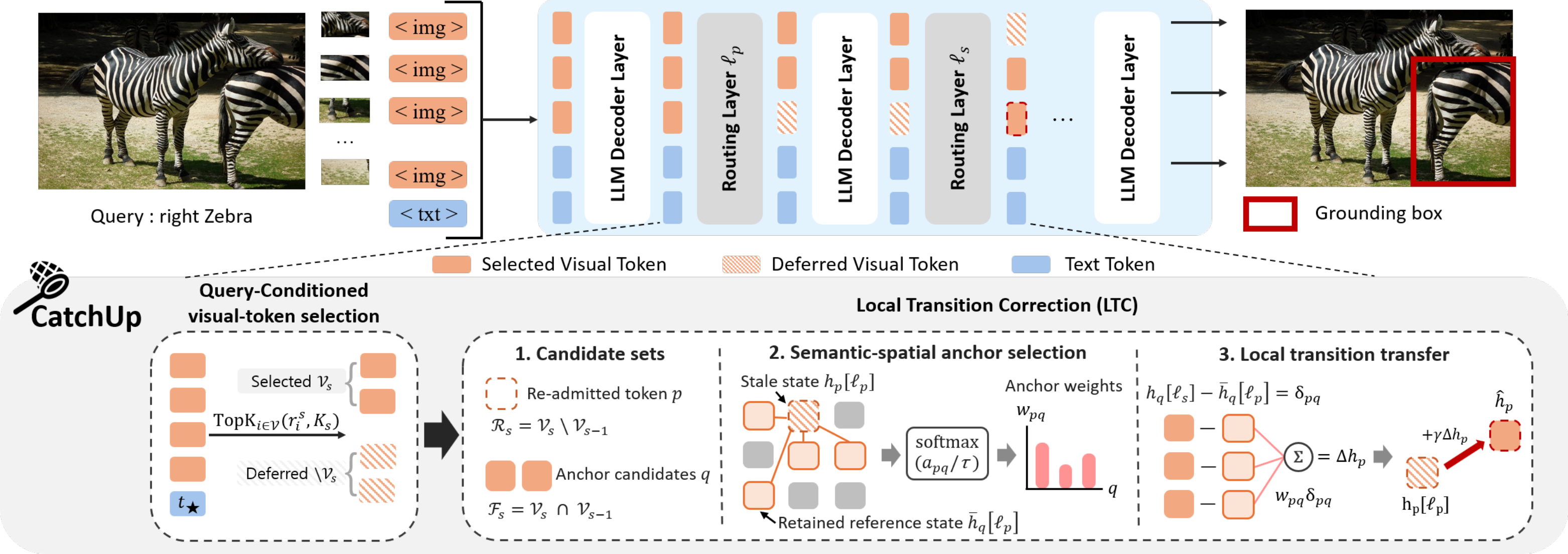}
    \caption{
    Overview of CatchUp.
    Reversible routing re-evaluates visual-token relevance across decoder layers, processing only selected tokens while preserving deferred tokens for later re-admission.
    LTC corrects newly re-admitted representations by transferring reference-to-current hidden-state changes from semantically and spatially related active anchors.
    }
    \label{fig:method_overview}
\end{figure*}


\subsection{Overview}
\label{sec:method_overview}

Motivated by \Cref{sec:observation_evidence,sec:observation_staleness},
we introduce \textbf{CatchUp}, a training-free framework combining reversible visual-token routing with \emph{Local Transition Correction} (LTC).
Reversible routing preserves deferred visual tokens for later reconsideration, while LTC approximates their skipped representation evolution upon re-admission.

Let $\mathcal V$ denote the visual-token index set and $L$ the total number of decoder layers, indexed by $\{0,\ldots,L-1\}$.
CatchUp performs visual-token routing at a subset of these layers,
\[
\mathcal S
=
\{\ell_1,\ell_2,\ldots,\ell_S\}
\subseteq
\{0,\ldots,L-1\},
\quad
\ell_1 < \ell_2 < \cdots < \ell_S,
\]
where $S=|\mathcal S|$ is the number of routing layers, $\ell_s$ denotes the $s$-th routing layer, and $\mathcal V_s\subseteq\mathcal V$ denotes the visual tokens selected at $\ell_s$.
We denote by $h_i[\ell]$ the representation of token $i$ available at decoder depth $\ell$.
Routing is performed at the input to layer $\ell_s$, before its attention and feed-forward computation, and all visual tokens remain active before the first routing layer $\ell_1$.

At each routing layer, CatchUp selects $\mathcal V_s$ from the complete visual-token set, allowing previously deferred tokens to return.
LTC is then applied only to newly re-admitted tokens before the selected tokens execute layer $\ell_s$.
Thus, LTC does not affect the routing decision at the same layer.
\Cref{fig:method_overview} summarizes the procedure.


\subsection{Reversible Visual Evidence Routing}
\label{sec:reversible_routing}

\paragraph{Query-conditioned visual-token selection.}
At routing layer $\ell_s$, CatchUp re-evaluates the
complete visual-token candidate set using the
current-layer text query and each token's latest
available representation.
Active tokens use their current-layer states,
whereas deferred tokens retain their latest cached
hidden states.
Rather than propagating deferred tokens through
skipped decoder layers, CatchUp projects their
cached states using the key projection of the
current routing layer and scores them against
the current-layer text query.

Let $\widetilde{\mathbf A}_{s}^{(m)}$ denote the
resulting post-softmax routing attention of head $m$,
and let $t_\star$ denote the last text-token position.
The relevance score of visual token $i\in\mathcal V$ is
\begin{equation}
r_i^s
=
\frac{1}{N_h}
\sum_{m=1}^{N_h}
\widetilde{\mathbf A}_{s}^{(m)}[t_\star,i],
\label{eq:routing_score}
\end{equation}
where $N_h$ denotes the number of attention heads.

Given the visual-token budget $K_s$ at routing
layer $\ell_s$, we select
\begin{equation}
\mathcal V_s
=
\operatorname{TopK}_{i\in\mathcal V}
\left(
r_i^s; K_s
\right).
\label{eq:routing_selection}
\end{equation}

Selection over the complete original token set
$\mathcal V$ allows excluded tokens to become
eligible for later reconsideration.
Although deferred tokens are scored using stale
states, \Cref{sec:observation_staleness} shows
that these states retain useful query-relevance
information for re-admission.

\paragraph{Reversible token execution.}
After selection at routing layer $\ell_s$, all text
tokens and the visual tokens in $\mathcal V_s$
are processed through the subsequent decoder interval.
For $s<S$, this interval spans decoder layers
$\ell_s,\ldots,\ell_{s+1}-1$, with the next routing
decision performed at the input to layer $\ell_{s+1}$.
After the final routing layer $\ell_S$, selected tokens
are processed through layers $\ell_S,\ldots,L-1$.
Visual tokens outside $\mathcal V_s$ are deferred;
their cached states remain unchanged until the
next routing decision.

For each deferred token, CatchUp tracks the decoder depth associated with its latest cached representation.
To support LTC, CatchUp retains the current visual-state snapshot and only those historical snapshots still referenced by deferred tokens, releasing snapshots once they are no longer needed.
Implementation details are provided in Appendix~A.

Before the first routing decision, all visual tokens
are active, and we define $\mathcal V_0=\mathcal V$.
The set of newly re-admitted tokens at routing
layer $\ell_s$ is
\begin{equation}
\mathcal R_s
=
\mathcal V_s
\setminus
\mathcal V_{s-1}.
\label{eq:readmitted_set}
\end{equation}


CatchUp applies LTC to tokens in $\mathcal R_s$ before they resume active decoder computation.


\subsection{Local Transition Correction}
\label{sec:ltc}

For a newly re-admitted token $p\in\mathcal R_s$, let $h_p[\ell_p]$ denote its cached representation at reference depth $\ell_p<\ell_s$.
Rather than recomputing the skipped decoder layers, LTC approximates the missing representation evolution using transitions from related active tokens.
For simplicity, we suppress the routing-step index $s$ for token-specific quantities.

\paragraph{Transition-anchor candidates.}
At routing layer $\ell_s$, we define the
candidate anchor set as
\begin{equation}
\mathcal F_s
=
\mathcal V_s
\cap
\mathcal V_{s-1}.
\label{eq:anchor_candidates}
\end{equation}

Since $\mathcal F_s\cap\mathcal R_s=\varnothing$,
a newly re-admitted token is never used as an
anchor at the same routing layer.
A re-admitted token can, however, become eligible
as an anchor at a subsequent routing layer if
it remains selected.

A token $q\in\mathcal F_s$ is selected at both
the previous and current routing layers and
has therefore been updated through the
immediately preceding decoder interval.
Its state $h_q[\ell_s]$ represents the token
at the current routing layer.

Let $\bar h_q[\ell]$ denote the retained reference
state of token $q$ from the visual-state snapshot
at layer $\ell$.
For a re-admitted token $p$ whose latest cached
representation is associated with reference depth
$\ell_p$, the pair
$(\bar h_q[\ell_p],h_q[\ell_s])$
provides a reference-to-current transition signal
that can be transferred to $p$.

\paragraph{Semantic--spatial anchor selection.}
Not all candidate anchors provide equally useful
transition signals for a re-admitted token.
Because a re-admitted token lacks an updated
current-depth representation, we measure
representation similarity using the retained
visual-state snapshot at its reference depth
$\ell_p$, providing a common reference for
the deferred token and candidate anchors.
For each candidate anchor $q\in\mathcal F_s$,
we compute
\begin{equation}
a_{pq}
=
\cos
\left(
h_p[\ell_p],
\bar h_q[\ell_p]
\right)
-
\lambda d_{\mathrm{sp}}(p,q),
\label{eq:anchor_score}
\end{equation}
where $\cos(\cdot,\cdot)$ denotes cosine similarity,
$d_{\mathrm{sp}}(p,q)$ denotes the Euclidean
distance between patch-grid coordinates after
normalizing each spatial axis to $[0,1]$,
and $\lambda$ controls the contribution of
spatial distance to anchor selection.
If $\mathcal F_s=\varnothing$, LTC is skipped
and the cached representation $h_p[\ell_p]$
is used unchanged.

Otherwise, we retain up to the $M$
highest-scoring anchors,
\begin{equation}
\mathcal N_p
=
\operatorname{TopM}_{q\in\mathcal F_s}
\left(
a_{pq}
\right),
\label{eq:anchor_selection}
\end{equation}
and compute their normalized anchor weights as
\begin{equation}
w_{pq}
=
\frac{
\exp\left(a_{pq}/\tau\right)
}{
\sum_{q'\in\mathcal N_p}
\exp\left(a_{pq'}/\tau\right)
},
\qquad
q\in\mathcal N_p,
\label{eq:anchor_weight}
\end{equation}
where $\tau$ is a temperature parameter.

\paragraph{Local transition transfer.}
For each selected anchor $q\in\mathcal N_p$,
its reference-to-current representation
displacement is
\begin{equation}
\delta_{pq}
=
h_q[\ell_s]
-
\bar h_q[\ell_p].
\label{eq:anchor_transition}
\end{equation}

LTC estimates the missing transition of
re-admitted token $p$ by aggregating
these local displacements:
\begin{equation}
\Delta h_p
=
\sum_{q\in\mathcal N_p}
w_{pq}\delta_{pq}.
\label{eq:transition_estimate}
\end{equation}

Rather than replacing the re-admitted token
with an anchor representation, LTC transfers
only the estimated representation change
and adds it to the token's own stale state:
\begin{equation}
\widehat h_p
=
h_p[\ell_p]
+
\gamma \Delta h_p,
\label{eq:ltc_correction}
\end{equation}
where $\gamma$ controls the correction strength.
This residual formulation retains the token's
own stale representation as the base state
while approximating the evolution missed
during deferral.
The corrected state $\widehat h_p$ then
serves as the representation of token $p$
when it resumes active computation.

Unlike the current-depth oracle in
\Cref{sec:observation_staleness},
LTC requires no full-token reference trajectory
during inference.
Instead, it approximates skipped representation
evolution using retained reference states and
transitions from active anchors.
We evaluate whether LTC improves alignment
with current-depth oracle representations
in \Cref{sec:alignment_analysis}.

\section{Experiments}
\label{sec:experiment}


\subsection{Experimental Setup}
\label{sec:experimental_setup}

\paragraph{Baselines.}
We compare CatchUp with FastV~\citep{chen2024fastv},
SparseVLM~\citep{zhang2025sparsevlm},
PyramidDrop~\citep{xing2025pyramiddrop},
N\"uwa~\citep{huang2026nuwa}, and the recoverable method
ReRoute~\citep{yang2026reroute} on
LLaVA-1.5-7B~\citep{liu2024llava15}.
The dense 576-token model serves as the full reference;
efficient methods are compared at matched routing budgets
where applicable.


\begin{table*}[t]
\centering
\caption{
Main results on eight referring-expression grounding splits
and six multimodal benchmarks.
CatchUp leads all grounding splits at every budget and
retains 92.17\% of dense-model multimodal performance at $K=64$.
Avg. Retention averages per-benchmark or per-split performance
ratios relative to full LLaVA-1.5-7B; bold denotes the best
efficient method at each budget.
}
\label{tab:main_results}

\scriptsize
\setlength{\tabcolsep}{2.0pt}
\renewcommand{\arraystretch}{0.92}
\setlength{\aboverulesep}{0.3ex}
\setlength{\belowrulesep}{0.3ex}

\resizebox{\textwidth}{!}{
\begin{tabular}{@{}l|ccc|ccc|cc|c||cccccc|c@{}}
\toprule

\textbf{Method}
& \multicolumn{9}{c||}{\textbf{Visual Grounding}}
& \multicolumn{7}{c}{\textbf{MLLM Benchmarks}} \\

\cmidrule(lr){2-10}
\cmidrule(lr){11-17}

&
\multicolumn{3}{c|}{\textbf{RefCOCO}}
& \multicolumn{3}{c|}{\textbf{RefCOCO+}}
& \multicolumn{2}{c|}{\textbf{RefCOCOg}}
& \textbf{Avg.}
& \textbf{POPE}
& \textbf{VQAv2}
& \textbf{VizWiz}
& \textbf{GQA}
& \textbf{MM-Vet}
& \textbf{MMBench}
& \textbf{Avg.} \\

&
\textbf{val} & \textbf{A} & \textbf{B}
& \textbf{val} & \textbf{A} & \textbf{B}
& \textbf{val} & \textbf{test}
& \textbf{Retention}
& & & & & & &
\textbf{Retention} \\

\midrule

\multicolumn{17}{c}{\textit{Dense Reference, 576 Tokens (100\%)}} \\

LLaVA-1.5-7B
& 54.15 & 59.27 & 47.97
& 46.53 & 53.93 & 39.05
& 46.89 & 47.67
& 100.00\%
& 85.90 & 75.84 & 54.30
& 60.47 & 28.67 & 62.37
& 100.00\% \\

\midrule

\multicolumn{17}{c}{\textit{Routing Budget $K=192$ ($\downarrow$67\%)}} \\

FastV {\scriptsize (ECCV'24)}
& 25.38 & 27.91 & 21.94
& 21.09 & 23.68 & 18.04
& 21.55 & 20.52
& 45.52\%
& 79.66 & 69.90 & 54.72
& 55.04 & 25.09 & 58.68
& 93.05\% \\

SparseVLM {\scriptsize (ICML'25)}
& 36.22 & 41.63 & 30.97
& 30.54 & 35.30 & 23.95
& 30.89 & 30.11
& 65.39\%
& 83.27 & 74.15 & 53.96
& 58.02 & 23.94 & 60.83
& 95.18\% \\

PyramidDrop {\scriptsize (CVPR'25)}
& 39.68 & 44.85 & 33.23
& 33.57 & 39.36 & 26.00
& 33.03 & 32.63
& 71.10\%
& \textbf{84.39} & 74.74 & 54.71
& 59.37 & 27.11 & 61.43
& 98.13\% \\

N\"uwa {\scriptsize (ICLR'26)}
& 18.23 & 23.44 & 14.29
& 14.81 & 19.65 & 11.37
& 14.85 & 14.43
& 32.79\%
& 84.06 & 73.50 & 55.31
& 57.66 & 27.16 & 56.01
& 96.09\% \\

ReRoute {\scriptsize (arXiv'26)}
& 45.28 & 49.81 & 39.27
& 38.13 & 44.22 & 31.01
& 39.43 & 39.21
& 82.40\%
& 84.11 & 74.87 & 54.56
& 58.94 & 26.28 & 61.77
& 97.55\% \\

\rowcolor{ourblue}
\textbf{CatchUp (Ours)}
& \textbf{47.23} & \textbf{51.55} & \textbf{41.90}
& \textbf{40.44} & \textbf{45.63} & \textbf{33.83}
& \textbf{40.24} & \textbf{40.63}
& \textbf{86.34\%}
& 84.31 & \textbf{75.19} & \textbf{55.47}
& \textbf{59.68} & \textbf{28.39} & \textbf{61.80}
& \textbf{99.37\%} \\

\midrule

\multicolumn{17}{c}{\textit{Routing Budget $K=128$ ($\downarrow$78\%)}} \\

FastV {\scriptsize (ECCV'24)}
& 10.85 & 12.71 & 9.13
& 8.63 & 10.29 & 7.85
& 9.27 & 9.13
& 19.65\%
& 72.96 & 63.94 & \textbf{54.61}
& 50.48 & 21.83 & 53.78
& 85.94\% \\

SparseVLM {\scriptsize (ICML'25)}
& 28.24 & 32.97 & 21.45
& 22.90 & 27.98 & 18.18
& 22.85 & 22.62
& 49.54\%
& 80.64 & 71.42 & 53.95
& 56.07 & 21.33 & \textbf{60.31}
& 91.87\% \\

PyramidDrop {\scriptsize (CVPR'25)}
& 30.90 & 36.29 & 24.04
& 25.26 & 30.54 & 18.43
& 24.42 & 24.16
& 53.66\%
& \textbf{81.98} & 72.81 & 53.98
& 56.91 & 23.17 & 59.97
& 93.66\% \\

N\"uwa {\scriptsize (ICLR'26)}
& 7.02 & 8.43 & 6.56
& 6.27 & 6.95 & 5.40
& 6.24 & 6.44
& 13.48\%
& 81.45 & 72.21 & 53.61
& 56.15 & 24.27 & 54.73
& 92.34\% \\

ReRoute {\scriptsize (arXiv'26)}
& 34.07 & 39.67 & 27.46
& 27.55 & 33.48 & 22.13
& 29.73 & 29.70
& 61.35\%
& 80.94 & 73.05 & 54.52
& 57.28 & 25.18 & 59.45
& 94.80\% \\

\rowcolor{ourblue}
\textbf{CatchUp (Ours)}
& \textbf{36.97} & \textbf{42.37} & \textbf{31.80}
& \textbf{29.77} & \textbf{36.34} & \textbf{24.30}
& \textbf{31.27} & \textbf{31.21}
& \textbf{66.48\%}
& 81.70 & \textbf{73.78} & 54.45
& \textbf{57.61} & \textbf{25.91} & \textbf{60.31}
& \textbf{95.84\%} \\

\midrule

\multicolumn{17}{c}{\textit{Routing Budget $K=64$ ($\downarrow$89\%)}} \\

FastV {\scriptsize (ECCV'24)}
& 2.50 & 0.16 & 0.53
& 0.33 & 0.07 & 0.51
& 0.10 & 0.08
& 1.06\%
& 51.07 & 43.25 & 51.02
& 40.00 & 12.75 & 22.42
& 59.50\% \\

SparseVLM {\scriptsize (ICML'25)}
& 4.00 & 4.26 & 2.98
& 3.24 & 3.51 & 2.35
& 3.03 & 3.08
& 6.65\%
& 66.31 & 58.63 & 52.82
& 48.58 & 18.76 & 45.88
& 78.52\% \\

PyramidDrop {\scriptsize (CVPR'25)}
& 12.95 & 14.71 & 10.09
& 10.18 & 12.35 & 8.24
& 10.38 & 10.22
& 22.40\%
& 74.81 & 66.63 & 52.51
& 52.68 & 18.90 & 55.07
& 85.50\% \\

N\"uwa {\scriptsize (ICLR'26)}
& 4.27 & 3.96 & 4.28
& 4.24 & 3.53 & 4.56
& 4.15 & 4.41
& 8.62\%
& 74.09 & 69.19 & 52.46
& 53.98 & 18.39 & 51.37
& 84.98\% \\

ReRoute {\scriptsize (arXiv'26)}
& 18.73 & 23.74 & 14.41
& 15.48 & 20.01 & 11.29
& 16.44 & 16.00
& 34.07\%
& 74.92 & 69.61 & 53.96
& 54.68 & 20.96 & 57.56
& 89.03\% \\

\rowcolor{ourblue}
\textbf{CatchUp (Ours)}
& \textbf{21.54} & \textbf{25.24} & \textbf{17.00}
& \textbf{17.02} & \textbf{20.73} & \textbf{13.15}
& \textbf{17.66} & \textbf{17.45}
& \textbf{37.60\%}
& \textbf{76.62} & \textbf{70.56} & \textbf{54.38}
& \textbf{55.16} & \textbf{24.95} & \textbf{57.62}
& \textbf{92.17\%} \\

\bottomrule
\end{tabular}
}
\end{table*}

\paragraph{Evaluation Benchmarks.}
We evaluate referring-expression grounding on
RefCOCO and RefCOCO+~\citep{yu2016modeling}
(val/TestA/TestB) and RefCOCOg~\citep{mao2016generation}
(val/test), reporting Acc@0.5.
For ablation and efficiency analyses, a single RefCOCO
score denotes the average over its three splits.
For general multimodal capability, we use
POPE~\citep{li2023pope}, VQAv2~\citep{goyal2017vqav2},
VizWiz~\citep{gurari2018vizwiz}, GQA~\citep{hudson2019gqa},
MM-Vet~\citep{yu2024mmvet}, and MMBench~\citep{liu2024mmbench}
with their standard metrics.

\paragraph{Experimental Settings.}
Unless otherwise specified, LLaVA-1.5-7B uses 576 initial
visual tokens and routing budgets $K\in\{64,128,192\}$.
We additionally evaluate Qwen2.5-VL-7B-Instruct~\citep{bai2025qwen25vl}
at average visual-token budgets of 64, 128, and 192.
CatchUp routes at decoder layers $\{2,7,15,23\}$ with
$M=8$, $\lambda=0.15$, $\gamma=0.8$, and $\tau=0.07$.
All hyperparameters are fixed across backbones, datasets, and budgets; pretrained parameters remain frozen. Further implementation details and hyperparameter sensitivity
analyses are provided in Appendices~A and C.8, respectively.


\subsection{Main Results}
\label{sec:main_results}

\paragraph{Referring-Expression Grounding.}
The consistent gains over ReRoute across routing budgets
indicate that preserving access to deferred visual evidence
alone is insufficient for accurate localization.
The sharp decline in grounding performance at smaller budgets,
even with recoverable routing, further suggests that fine-grained
localization is sensitive to both evidence selection and the
representation quality of returning tokens.
The contribution of correction is isolated in
\Cref{sec:alignment_analysis}.
On Qwen2.5-VL-7B-Instruct, CatchUp also achieves the best
RefCOCO accuracy among evaluated efficient methods at all
three budgets, supporting its grounding benefits beyond
the LLaVA-1.5 backbone.

\paragraph{General Multimodal Benchmarks.}
General multimodal performance is markedly more resilient to token reduction than fine-grained grounding: at $K=64$, CatchUp retains 92.17\% of dense performance across six benchmarks, versus 37.60\% for grounding (\Cref{tab:main_results}).
This disparity suggests that precise spatial localization requires visual evidence that is harder to preserve under aggressive token budgets.
On Qwen2.5-VL-7B-Instruct, CatchUp achieves the best or tied-best POPE and VQAv2 results in four of six budget--benchmark comparisons, with gaps of at most
0.07 points in the remaining two.
Full additional-backbone results are provided in Appendix~C.7.


\begin{figure}[t]
    \centering
    \includegraphics[width=0.98\columnwidth]
    {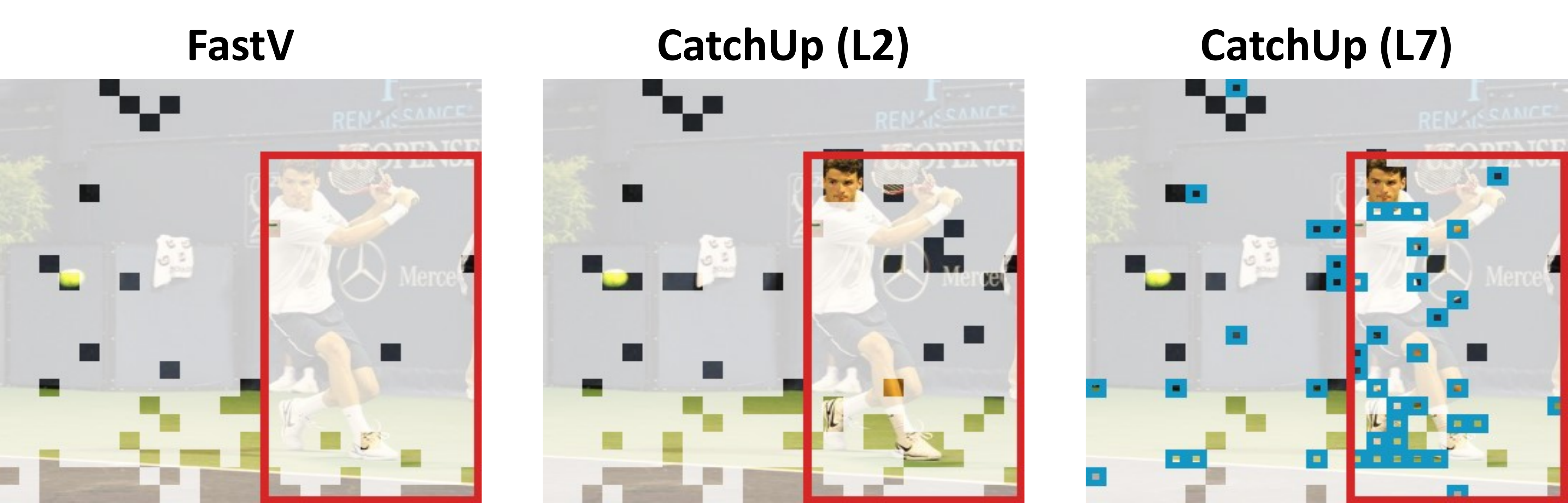}
    \caption{
    Reversible visual-token routing for the referring
    expression \emph{player}.
    FastV shows the final tokens after irreversible
    pruning, while CatchUp shows selected tokens
    at L2 and L7, including newly re-admitted tokens.
    The red box marks the ground-truth target;
    whitened regions denote deferred tokens,
    and blue outlines mark tokens re-admitted at L7.
    }
    \label{fig:qualitative}
\end{figure}

\paragraph{Qualitative Analysis.}
\Cref{fig:qualitative} illustrates late re-entry of
target-region tokens excluded by early selection,
which irreversible pruning cannot recover.
This visualizes evidence recoverability; the separate
question of representation readiness is examined in
\Cref{sec:alignment_analysis}.


\subsection{Re-Admission and Alignment Analysis}
\label{sec:alignment_analysis}

\begin{table}[t]
\centering
\caption{
Component ablation at $K=64$.
Reversible re-admission improves RefCOCO from 15.78 to 18.96;
LTC further raises it to 21.26, outperforming
absolute-state mixing.
}
\label{tab:component_ablation}

\scriptsize
\setlength{\tabcolsep}{1.5pt}
\renewcommand{\arraystretch}{0.98}

\begin{adjustbox}{width=\columnwidth}
\begin{tabular}{llrrr}
\toprule

\textbf{Variant}
& \textbf{Treatment}
& \textbf{RefCOCO} $\uparrow$
& \textbf{POPE} $\uparrow$
& \textbf{VQAv2} $\uparrow$ \\

\midrule

Monotonic Compact Routing
& --
& 15.78 & 75.81 & 67.75 \\

CatchUp w/o LTC
& Stale
& 18.96 & 74.92 & 69.61 \\

CatchUp + Absolute
& Absolute-state mixing
& 19.31 & 75.20 & 69.50 \\

\rowcolor{ourblue}
\textbf{CatchUp}
& \textbf{Transition residual}
& \textbf{21.26} & \textbf{76.62} & \textbf{70.56} \\

\bottomrule
\end{tabular}
\end{adjustbox}
\end{table}

\paragraph{Contribution of Re-Admission and Correction.}
\Cref{tab:component_ablation} distinguishes two
complementary benefits: reversible routing restores
access to deferred evidence, while LTC makes
re-admitted states more suitable for computation
at the current depth.
The limited gain from absolute-state mixing (defined in Appendix~C.2), despite using related anchors, indicates that transferring how representations change is more effective than blending their absolute states.

\begin{table}[t]
\centering
\caption{
Representation alignment with current-depth oracle states
on 500 RefCOCO validation samples ($K=64$).
The Events column counts individual re-admitted visual tokens
across samples and routing layers; counts may differ because
correction can alter subsequent routing decisions.
Gap-wise columns report cosine similarity.
}
\label{tab:representation_alignment}

\scriptsize
\setlength{\tabcolsep}{2.5pt}
\renewcommand{\arraystretch}{1.10}

\begin{adjustbox}{max width=\columnwidth}
\begin{tabular}{@{}lrrr@{\hspace{7pt}}rrr@{}}
\toprule

& \multicolumn{3}{c}{\textbf{Overall Alignment}}
& \multicolumn{3}{c}{\textbf{Cosine by Gap}} \\

\cmidrule(lr){2-4}
\cmidrule(l){5-7}

\textbf{Variant}
& \textbf{Events}
& \textbf{Cos.} $\uparrow$
& \textbf{Norm. $\ell_2$} $\downarrow$
& \textbf{5--8}
& \textbf{9--12}
& \textbf{13+} \\

\midrule

Stale
& 37,533 & 0.6299 & 0.8557
& 0.8066 & 0.4926 & 0.3947 \\

Absolute-state mixing
& 36,573 & 0.6241 & 0.9513
& 0.8342 & 0.4237 & 0.3326 \\

\rowcolor{ourblue}
\textbf{LTC}
& 36,959
& \textbf{0.7373}
& \textbf{0.6893}
& \textbf{0.9132}
& \textbf{0.5949}
& \textbf{0.4944} \\

\bottomrule
\end{tabular}
\end{adjustbox}
\end{table}

\paragraph{Representation Alignment.}
\Cref{tab:representation_alignment} shows that LTC improves both directional and normalized-distance agreement with the same-token current-depth oracle,
whereas absolute-state mixing does not.
Since corrections can alter later routing trajectories, the overall event sets differ across variants.
Crucially, LTC also outperforms absolute-state mixing on sample-, layer-, and token-matched re-admission events, supporting the benefit of transition transfer
independently of differences in event composition (see Appendix~C.3).

\paragraph{Effect of Re-Admission Gap.}
The gap-wise results in \Cref{tab:representation_alignment}
reveal that stale states become increasingly misaligned
as the re-admission gap grows.
LTC improves directional alignment in every gap range,
but its remaining long-gap error reflects the difficulty
of approximating accumulated nonlinear decoder
transformations with local transitions.
Gap-wise normalized $\ell_2$ results are reported in Appendix~C.4.

\paragraph{LTC Anchor Design.}
Semantic and spatial cues are complementary across tasks rather than uniformly optimal for every metric.
Their combination leads on three of four evaluated benchmarks, whereas spatial-only anchors slightly favor RefCOCO, supporting the combined design
as a balanced default.
Full results are provided in Appendix~C.5.


\subsection{Efficiency Analysis}
\label{sec:efficiency}

\begin{figure}[t]
    \centering
    \includegraphics[width=\columnwidth]
    {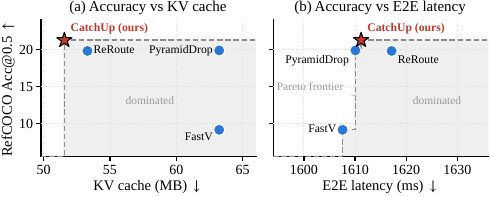}
    \caption{
    RefCOCO accuracy versus (a) decoder KV-cache memory
    and (b) end-to-end latency on LLaVA-1.5-7B
    at matched prefill TFLOPs (within 1\%).
    CatchUp lies on the non-dominated accuracy--efficiency
    frontier.
    Dashed lines mark the Pareto frontier;
    shaded regions are dominated.
    Dense inference requires 8.673 TFLOPs.
    }
    \label{fig:efficiency_tradeoff}
\end{figure}

\paragraph{Efficiency--Accuracy Trade-off.}
\Cref{fig:efficiency_tradeoff} shows that CatchUp
achieves higher grounding accuracy at matched prefill
computation while maintaining a small decoder KV cache.
However, KV-cache memory excludes LTC reference snapshots
and should not be interpreted as total GPU-memory savings.
At $K=64$, prefill latency decreases by 55.3\%
relative to dense inference, whereas end-to-end latency
decreases by only 8.0\%, reflecting the dominance
of autoregressive decoding.

\begin{table}[t]
\centering
\caption{
LTC overhead at $K=64$.
Transition correction improves RefCOCO by 2.30 points
with 2.96\% additional prefill TFLOPs,
1.38\% higher end-to-end latency, and
21.2\,MiB additional peak PyTorch CUDA allocation.
}
\label{tab:ltc_overhead}

\scriptsize
\setlength{\tabcolsep}{2pt}
\renewcommand{\arraystretch}{1.12}

\resizebox{\columnwidth}{!}{
\begin{tabular}{@{}lrrrr@{}}
\toprule

\textbf{Variant}
& \textbf{RefCOCO} $\uparrow$
& \textbf{TFLOPs} $\downarrow$
& \textbf{E2E (ms)} $\downarrow$
& \textbf{Peak CUDA (MiB)} $\downarrow$ \\

\midrule

w/o LTC
& 18.96
& 2.131
& 1589.28
& 17799.3 \\

\rowcolor{ourblue}
\textbf{CatchUp}
& \textbf{21.26}
& 2.194
& 1611.17
& 17820.5 \\

\bottomrule
\end{tabular}
}
\end{table}

\paragraph{LTC Overhead.}
\Cref{tab:ltc_overhead} shows that LTC improves grounding accuracy with modest additional computation, latency, and peak memory allocation, without
re-executing the skipped decoder layers.
Despite this overhead, CatchUp's peak CUDA allocation remains 602.2\,MiB (3.3\%) below dense inference.
Detailed latency and memory breakdowns, including LTC reference snapshots, are provided in Appendices~D.2 and D.3.



\section{Conclusion}
\label{sec:conclusion}

We presented \textbf{CatchUp}, a training-free framework combining reversible visual-token routing with Local Transition Correction (LTC) to address representation staleness upon re-admission.
Controlled oracle interventions reveal the impact of representation staleness, while LTC improves current-depth alignment without recomputing skipped decoder layers.
CatchUp achieves the best accuracy among evaluated efficient methods across all eight grounding splits at 64-, 128-, and 192-token budgets.
At $K=64$, it retains 92.17\% of dense-model performance across six multimodal benchmarks while reducing end-to-end latency by 8.0\%.
These findings highlight the importance of both evidence recoverability and representation readiness.

\section*{Acknowledgement}
This work was supported by the National Research Foundation of Korea (NRF) grant funded by the Korea government (MSIT) (RS-2025-00561169), Global – Learning \& Academic research institution for Master’s · PhD students, and Postdocs (G-LAMP) Program of the National Research Foundation of Korea (NRF) grant funded by the Ministry of Education (No. RS-2025-25442252), Institute of Information \& communications Technology Planning \& Evaluation (IITP) under the Leading Generative AI Human Resources Development (IITP-2027-RS-2026-25546026) grant funded by the Korea government (MSIT), and Korea Internet \& Security Agency(KISA) grant funded by the Korea government(PIPC) (No.2780000041, Development of PET-based Generation and Validation Technology for Unstructured Time-Series Synthetic Data).

{
    \small
    \bibliographystyle{ieeenat_fullname}
    \bibliography{main}
}

\clearpage

\setcounter{table}{6}
\setcounter{figure}{6}
\setcounter{equation}{12}

\appendix


\section{Implementation Details}
\label{sec:supp_impl}

We provide additional implementation details of CatchUp,
including persistent token states, reference-snapshot
management, and LTC execution.
All CatchUp operations are performed during inference
without updating pretrained model parameters.

\subsection{Reversible Routing and Persistent Token States}
\label{sec:supp_routing}

CatchUp maintains a persistent logical entry for every
original visual token throughout inference.
Each entry stores the token's original index, patch-grid
position, latest hidden representation, and the decoder
depth associated with that representation.
Routing changes whether a token participates in decoder
computation without discarding its identity, spatial
correspondence, or latest available state.

At routing layer $\ell_s$, CatchUp reconstructs the
complete candidate set and computes the query-conditioned
relevance scores defined in the main paper.
The selected visual-token set is
\begin{equation}
\mathcal V_s
=
\operatorname{TopK}_{i\in\mathcal V}
(r_i^s;K_s),
\end{equation}
where $\mathcal V$ is the complete set of original
visual tokens.
Active tokens are scored using their current-layer
representations, whereas deferred tokens use their
latest cached states.
For deferred tokens, routing keys are recomputed
with the current layer's key projection rather than
reusing keys from earlier layers.
Routing occurs at the input to decoder layer $\ell_s$,
before that layer is executed.

After selection and any required LTC, the text tokens
and selected visual tokens are gathered into a compact
sequence, preserving the original relative order of
selected visual tokens.
For $s<S$, this sequence is processed through layers
$\ell_s,\ldots,\ell_{s+1}-1$, followed by another
routing decision at the input to layer $\ell_{s+1}$.
After the final routing layer, selected tokens are
processed through layers $\ell_S,\ldots,L-1$.

Updated visual-token states are scattered back to
their corresponding persistent entries, and their
associated reference depths are updated.
Deferred tokens bypass the interval, leaving both
their cached states and reference depths unchanged.
Their persistent entries remain eligible for scoring
at subsequent routing layers.

Before the first routing decision, all visual tokens
are active, and we define $\mathcal V_0=\mathcal V$.
Newly re-admitted tokens at routing layer $\ell_s$ are
\begin{equation}
\mathcal R_s
=
\mathcal V_s
\setminus
\mathcal V_{s-1}.
\end{equation}
Thus, $\mathcal R_1=\varnothing$, since no visual token
has been deferred before the first routing decision.

\subsection{Need-Based Reference-State Management}
\label{sec:supp_reference}

For each visual token $i$, CatchUp tracks a reference
depth $\ell_i$ associated with its latest cached state.
When a token remains deferred across multiple routing
intervals, its cached representation and reference
depth remain unchanged.
Consequently, a token $p$ re-admitted at layer $\ell_s$
may retain a representation from an earlier depth,
\begin{equation}
\ell_p < \ell_s.
\end{equation}

LTC requires historical anchor representations at
the reference depth $\ell_p$.
Rather than retaining the complete decoder history,
CatchUp uses need-based reference-snapshot management.
Each snapshot stores visual representations indexed
by their original token positions.
Following the main-paper notation,
$\bar h_q[\ell]$ denotes the retained reference state
of token $q$ associated with depth $\ell$.

Snapshots reflect the visual states associated with
their respective decoder depths.
At a routing layer, newly re-admitted tokens are
stored after LTC has been applied, so their corrected
representations become the corresponding reference
states.

A historical snapshot is retained while at least
one currently deferred token references its depth.
Once no deferred token requires that reference depth,
the snapshot is released.
The current visual-state snapshot is also retained
as a potential reference for tokens deferred at
the current routing decision.

Thus, snapshot retention follows the reference
depths of deferred tokens rather than a fixed
decoder-history window.
This avoids storing states from every decoder
layer and releases obsolete snapshots when they
are no longer needed.
Reference snapshots are used only for LTC and
do not add tokens to decoder computation or
the decoder KV cache.

If a required reference snapshot is unavailable,
LTC is skipped for the affected token and its
cached state is used unchanged.

\subsection{LTC Eligibility and Transition Anchors}
\label{sec:supp_ltc_eligibility}

LTC is applied only to newly re-admitted tokens
$p\in\mathcal R_s$.
We define the re-admission gap as
\begin{equation}
g_p
=
\ell_s-\ell_p,
\end{equation}
where $\ell_p$ is the reference depth associated
with the token's latest cached representation.

The transition-anchor candidate set is
\begin{equation}
\mathcal F_s
=
\mathcal V_s
\cap
\mathcal V_{s-1}.
\end{equation}

Since $\mathcal F_s\cap\mathcal R_s=\varnothing$,
newly re-admitted tokens cannot serve as anchors
at the same routing layer.
A previously re-admitted token may become an anchor
at a later routing layer if it remains selected.

Each $q\in\mathcal F_s$ is selected at consecutive
routing stages and has undergone the immediately
preceding decoder interval.
Its current representation can therefore provide
a transition signal for a returning token.

When fewer than $M$ valid anchor candidates are
available, all available candidates are used.
If the required reference snapshot is missing,
LTC is skipped and the cached representation of
the affected token remains unchanged.

\subsection{Semantic--Spatial Anchor Selection}
\label{sec:supp_anchor_impl}

For LLaVA-1.5, the 576 visual tokens correspond
to a $24\times24$ patch grid.
Let $(y_i,x_i)$ denote the coordinates of
visual token $i$, where
\begin{equation}
y_i,x_i\in\{0,\ldots,23\}.
\end{equation}

We normalize the spatial axes independently
to $[0,1]$:
\begin{equation}
\mathbf p_i
=
\left(
\frac{y_i}{23},
\frac{x_i}{23}
\right).
\end{equation}

The spatial distance between a re-admitted token
$p$ and anchor candidate $q$ is
\begin{equation}
d_{\mathrm{sp}}(p,q)
=
\left\|
\mathbf p_p-\mathbf p_q
\right\|_2.
\end{equation}
The maximum displacement along either axis is $1$,
and the maximum diagonal distance is $\sqrt{2}$.

Semantic similarity is evaluated at the reference
depth $\ell_p$ of the re-admitted token.
Its cached state $h_p[\ell_p]$ is compared
with the stored candidate state $\bar h_q[\ell_p]$.
Using a common reference depth avoids directly
comparing the deferred token's stale state with
an anchor's current-depth representation.

Candidates are ranked using the semantic--spatial
score defined in the main paper:
\begin{equation}
a_{pq}
=
\cos
\left(
h_p[\ell_p],
\bar h_q[\ell_p]
\right)
-
\lambda d_{\mathrm{sp}}(p,q).
\end{equation}

Up to the $M$ highest-scoring candidates are retained.
Their scores are converted into normalized aggregation
weights through the softmax with temperature $\tau$,
as defined in the main method.

\subsection{Transition Computation and State Update}
\label{sec:supp_transition_impl}

For each selected anchor $q\in\mathcal N_p$,
CatchUp computes its reference-to-current displacement:
\begin{equation}
\delta_{pq}
=
h_q[\ell_s]
-
\bar h_q[\ell_p].
\end{equation}

The missing transition of re-admitted token $p$
is estimated as
\begin{equation}
\Delta h_p
=
\sum_{q\in\mathcal N_p}
w_{pq}
\left(
h_q[\ell_s]
-
\bar h_q[\ell_p]
\right),
\end{equation}
where $w_{pq}$ is the normalized semantic--spatial
anchor weight.

LTC applies this transition to the token's own
cached representation:
\begin{equation}
\widehat h_p
=
h_p[\ell_p]
+
\gamma \Delta h_p.
\end{equation}

This update transfers local representation changes
rather than directly mixing absolute current-depth
anchor states.
The re-admitted token retains its own representation
as the base, while anchor transitions approximate
the evolution missed during deferral.

The corrected state $\widehat h_p$ replaces the
stale cached state and is used when token $p$
resumes active decoder computation.

\subsection{Fixed Implementation Configuration}
\label{sec:supp_configuration}

CatchUp uses a fixed configuration across the
evaluated datasets, backbones, and visual-token
budgets, without benchmark-specific or sample-specific
hyperparameter tuning.
Pretrained model parameters remain frozen.
The configuration is summarized in
\Cref{tab:supp_configuration}.

\begin{table}[t]
    \centering
    \small
    \setlength{\tabcolsep}{6pt}
    \renewcommand{\arraystretch}{1.08}
    \begin{tabular}{@{}lc@{}}
        \toprule
        \textbf{Parameter}
        & \textbf{Setting} \\
        \midrule
        Routing layers
        & $\{2,7,15,23\}$ \\
        Number of anchors $M$
        & 8 \\
        Spatial weight $\lambda$
        & 0.15 \\
        Temperature $\tau$
        & 0.07 \\
        Transition scale $\gamma$
        & 0.80 \\
        Snapshot retention
        & Need-based \\
        \bottomrule
    \end{tabular}
    \caption{
    Fixed CatchUp implementation configuration shared across evaluated datasets, backbones, and routing budgets.
    Hyperparameter sensitivity is evaluated in ~\Cref{sec:supp_hyperparameter_sensitivity}.
    }
    \label{tab:supp_configuration}
\end{table}


\section{CatchUp Inference Algorithm}
\label{sec:supp_algorithm}

\Cref{alg:catchup} summarizes CatchUp inference.
Persistent token-state management, reference-snapshot
retention, and anchor construction are described
in \Cref{sec:supp_impl}.

\begin{algorithm}[t]
\caption{CatchUp Inference}
\label{alg:catchup}

\begin{algorithmic}[1]

\Require
Visual tokens $\mathcal V$ and text tokens $\mathcal T$
at the first routing layer $\ell_1$;
routing layers $\{\ell_s\}_{s=1}^{S}$;
token budgets $\{K_s\}_{s=1}^{S}$

\State Initialize persistent states for all $i\in\mathcal V$
\State $\mathcal V_0 \gets \mathcal V$

\For{$s=1,\ldots,S$}

    \State Compute relevance scores
    $\{r_i^s\}_{i\in\mathcal V}$
    over the complete visual-token set

    \State
    $\mathcal V_s
    \gets
    \operatorname{TopK}_{i\in\mathcal V}
    (r_i^s;K_s)$

    \State
    $\mathcal R_s
    \gets
    \mathcal V_s\setminus\mathcal V_{s-1}$

    \ForAll{$p\in\mathcal R_s$}

        \State Construct valid anchors $\mathcal N_p$
        and weights $\{w_{pq}\}_{q\in\mathcal N_p}$

        \If{$\mathcal N_p\neq\varnothing$}

            \State
            $\displaystyle
            \Delta h_p
            \gets
            \sum_{q\in\mathcal N_p}
            w_{pq}
            \bigl(
            h_q[\ell_s]-\bar h_q[\ell_p]
            \bigr)$

            \State
            $\widehat h_p
            \gets
            h_p[\ell_p]+\gamma\Delta h_p$

        \Else

            \State
            $\widehat h_p\gets h_p[\ell_p]$

        \EndIf

        \State Replace the persistent state of $p$
        with $\widehat h_p$

    \EndFor

    \State Retain reference snapshots required by
    deferred tokens and release obsolete snapshots

    \State Gather $\mathcal T$ and $\mathcal V_s$
    into a compact sequence

    \State Process the compact sequence through
    the subsequent decoder interval

    \State Update persistent states and reference
    depths of selected visual tokens

\EndFor

\State \Return model output

\end{algorithmic}
\end{algorithm}


\section{Additional Experiments}
\label{sec:supp_exp}

This section provides experimental protocols,
ablation definitions, and additional analyses
supporting the main results.
Unless otherwise specified, ablation and
representation analyses use LLaVA-1.5-7B
with a 64-token routing budget and the fixed
CatchUp configuration.

\paragraph{Average retention.}
For the main results, average retention is
computed as
\begin{equation}
\mathrm{Avg.~Retention}
=
\frac{1}{N}
\sum_{i=1}^{N}
\frac{s_i}{s_i^{\mathrm{Full}}}
\times 100,
\end{equation}
where $s_i$ and $s_i^{\mathrm{Full}}$ are the
performance scores of the evaluated method
and the dense model, respectively.
We use $N=8$ for grounding splits and
$N=6$ for multimodal benchmarks.


\subsection{Additional Details for Motivating Observations}
\label{sec:supp_observation_details}

We detail the evaluation protocols and metrics
used for the motivating observations.

\paragraph{Observation I protocol.}
Evidence dynamics are analyzed on 500 RefCOCO
samples using LLaVA-1.5-7B under full-token
inference with 576 visual tokens.
At each decoder layer, relevance is measured
using post-softmax attention from the last
text token to each visual token, averaged
over attention heads.
We set $K=64$ and use decoder layer 2,
the first routing layer in our configuration,
as the reference depth.

Relevance volatility and the late-emerging
evidence ratio (LER) follow the definitions
in the main paper.
For reference, two independently sampled
top-64 subsets from 576 tokens have an
expected LER of
$1-64/576=88.89\%$.

\paragraph{Stale--oracle routing agreement.}
For the Observation II routing diagnostic,
stale and current-depth oracle routing use
the same current-layer text query and
the same current-layer key projection.
They differ only in the visual-token state
used to construct each key.

Let $\ell_i^s$ denote the reference depth
of token $i$ before routing at $\ell_s$.
The two relevance scores are
\begin{equation}
r_{i,\mathrm{stale}}^s
=
\operatorname{Attn}
\left(
q^{\ell_s},
K_{\ell_s}(h_i[\ell_i^s])
\right),
\end{equation}
and
\begin{equation}
r_{i,\mathrm{oracle}}^s
=
\operatorname{Attn}
\left(
q^{\ell_s},
K_{\ell_s}(h_{i,\mathrm{full}}[\ell_s])
\right).
\end{equation}

Here, $\operatorname{Attn}$ represents the
attention-based relevance aggregation defined
in the main paper.
The stale representation $h_i[\ell_i^s]$
is the token's latest available state, whereas
$h_{i,\mathrm{full}}[\ell_s]$ is the current-depth
state of the same token from a separate full-token
reference trajectory.
Oracle scores are used only for diagnostic
comparison and never change the actual
stale-routing trajectory.

For sample $n$, let
$\mathcal V_{\mathrm{stale},n}^s$ and
$\mathcal V_{\mathrm{oracle},n}^s$
denote the corresponding top-$K_s$ selections.
Overall Top-$K$ Overlap is micro-averaged
over sample--routing-stage events:
\begin{equation}
\operatorname{Overlap@K}
=
\frac{
\sum_{n,s}
\left|
\mathcal V_{\mathrm{stale},n}^s
\cap
\mathcal V_{\mathrm{oracle},n}^s
\right|
}{
\sum_{n,s}K_s
}.
\end{equation}

Let
\begin{equation}
\mathcal D_n^s
=
\mathcal V
\setminus
\mathcal V_{\mathrm{stale},n}^{s-1}
\end{equation}
denote the tokens deferred by the
stale-routing trajectory before stage $s$.

The stale and oracle re-admission sets are
\begin{equation}
\mathcal R_{\mathrm{stale},n}^s
=
\mathcal V_{\mathrm{stale},n}^s
\cap
\mathcal D_n^s,
\qquad
\mathcal R_{\mathrm{oracle},n}^s
=
\mathcal V_{\mathrm{oracle},n}^s
\cap
\mathcal D_n^s.
\end{equation}

Re-admission Recall@K is the fraction of
oracle-selected deferred tokens also selected
by stale routing:
\begin{equation}
\operatorname{ReAdmRecall@K}
=
\frac{
\sum_{n,s}
\left|
\mathcal R_{\mathrm{stale},n}^s
\cap
\mathcal R_{\mathrm{oracle},n}^s
\right|
}{
\sum_{n,s}
\left|
\mathcal R_{\mathrm{oracle},n}^s
\right|
}.
\end{equation}

Both metrics exclude the first routing stage,
where no visual token has yet been deferred.
Re-admission Recall@K is also micro-averaged
over sample--routing-stage events.

For the random reference, we uniformly sample
the same number of visual tokens as the
corresponding stale-routing selection and
compute the same agreement metrics.
Mean and standard deviation are reported
over five random seeds.

\paragraph{Oracle-state intervention.}
To isolate representation staleness after
re-admission, we replay a fixed routing trajectory,
holding the selected and deferred token sets
and visual-token budget unchanged at every
routing layer.

Only newly re-admitted token states are modified.
For each $j\in\mathcal R_s$, the stale condition
retains its cached state, whereas the oracle
condition replaces it with
$h_{j,\mathrm{full}}[\ell_s]$
from a separate full-token trajectory.
All other token states remain unchanged
at each intervention point.

The full-token reference trajectory is used
only for this diagnostic intervention and
is not required by CatchUp inference.


\subsection{Ablation Variant Definitions}
\label{sec:supp_ablation_variants}

We define the component-ablation and anchor-selection
variants used in the main paper and supplementary
experiments.
Unless stated otherwise, variants share the
routing configuration and token budget and
differ only in the ablated component.

\paragraph{Component-ablation variants.}
\emph{Monotonic Compact Routing} uses compact
visual-token execution without re-admission:
once excluded, a visual token remains unavailable
at subsequent routing layers.

\emph{CatchUp w/o LTC} permits reversible
selection over the complete visual-token set,
but re-admitted tokens resume computation
directly from their stale cached states.

\emph{Absolute-State Mixing} uses the same
reversible routing and semantic--spatial
anchor construction as CatchUp, but directly
mixes current-depth anchor states instead
of transferring representation transitions.

For re-admitted token $p$, we first aggregate
its anchors' current representations:
\begin{equation}
h_p^{\mathrm{anchor}}
=
\sum_{q\in\mathcal N_p}
w_{pq}h_q[\ell_s].
\end{equation}

The re-admitted representation is updated as
\begin{equation}
\widehat h_p
=
(1-\beta)h_p[\ell_p]
+
\beta h_p^{\mathrm{anchor}},
\end{equation}
where $\beta=0.2$.

Unlike Absolute-State Mixing, LTC transfers
only reference-to-current representation changes
(\Cref{sec:supp_transition_impl}).
This comparison tests whether correction is
better achieved through absolute anchor states
or the transitions those anchors undergo.

\paragraph{Anchor-selection variants.}
The anchor-selection ablation keeps the LTC
transition-residual update fixed and varies
only the rule for selecting anchors from
$\mathcal F_s$.

\emph{Random} uniformly samples up to $M$
valid candidates and uses uniform weights.

\emph{Spatial} ranks candidates using
normalized patch-grid proximity:
\begin{equation}
a_{pq}^{\mathrm{sp}}
=
-d_{\mathrm{sp}}(p,q).
\end{equation}

\emph{Semantic} uses representation similarity
at the re-admitted token's reference depth:
\begin{equation}
a_{pq}^{\mathrm{sem}}
=
\cos
\left(
h_p[\ell_p],
\bar h_q[\ell_p]
\right).
\end{equation}

\emph{Semantic + Spatial}, the default
CatchUp configuration, combines both:
\begin{equation}
a_{pq}^{\mathrm{sem+sp}}
=
\cos
\left(
h_p[\ell_p],
\bar h_q[\ell_p]
\right)
-
\lambda d_{\mathrm{sp}}(p,q).
\end{equation}

For score-based variants, the top-$M$
candidates are retained and their weights
are obtained using the same softmax
aggregation rule as the main method.


\subsection{Additional Representation-Alignment Details}
\label{sec:supp_representation}

The main paper reports overall alignment with
current-depth oracle states and cosine similarity
across re-admission gaps.
Here, we define the alignment metrics, examine
matched re-admission events, and report
complementary gap-wise normalized distances.

Unless stated otherwise, the analysis uses
500 common RefCOCO validation samples,
LLaVA-1.5-7B, and $K=64$.
Full-token trajectories are used only for
diagnosis, not during CatchUp inference.

\paragraph{Representation-alignment metrics.}
For a token $p$ re-admitted at layer $\ell_s$,
let $h_{p,\mathrm{full}}[\ell_s]$ be its
current-depth state from a separate
full-token inference trajectory.

Given the re-admitted representation
$\widehat h_p[\ell_s]$, directional agreement
with the oracle is measured by
\begin{equation}
\operatorname{CosSim}
=
\cos
\bigl(
\widehat h_p[\ell_s],
h_{p,\mathrm{full}}[\ell_s]
\bigr),
\end{equation}
and relative discrepancy by
\begin{equation}
\operatorname{NormL2}
=
\frac{
\left\|
\widehat h_p[\ell_s]
-
h_{p,\mathrm{full}}[\ell_s]
\right\|_2
}{
\left\|
h_{p,\mathrm{full}}[\ell_s]
\right\|_2+\epsilon
},
\end{equation}
where $\epsilon$ is a small constant
for numerical stability.

Higher cosine similarity and lower normalized
$\ell_2$ distance indicate better agreement
with the current-depth oracle.
We compare stale re-admission,
Absolute-State Mixing, and LTC.

\paragraph{Matched re-admission events.}
Representation correction may alter later routing
decisions, so variants can generate different
sets of re-admission events.
The main paper reports $37{,}533$ stale events,
$36{,}573$ Absolute-State Mixing events,
and $36{,}959$ LTC events.

To control for this difference, we additionally
compare LTC and Absolute-State Mixing on events
matched by sample, routing layer, and original
visual-token index.

Across $35{,}512$ matched events, LTC achieves
higher cosine similarity in $99.38\%$ of cases
and lower normalized $\ell_2$ distance in
$93.10\%$ of cases.
These results support the advantage of
transition transfer independently of
differences in event composition.


\subsection{Additional Analysis by Re-Admission Gap}
\label{sec:supp_gap}

For a token re-admitted at layer $\ell_s$
with reference depth $\ell_p$, the
re-admission gap is $g_p=\ell_s-\ell_p$.
Larger gaps indicate more skipped decoder
computation.
For example, a token associated with depth $6$
that returns at layer $15$ has $g_p=9$.

The main paper reports gap-wise cosine similarity.
\Cref{tab:supp_gap_l2} complements that analysis
with normalized $\ell_2$ discrepancy.

\begin{table}[t]
\centering
\caption{
Normalized $\ell_2$ discrepancy from current-depth
oracle states by re-admission gap ($K=64$).
LTC yields the lowest discrepancy in all three
ranges, with a smaller margin for gaps of 13+
layers. Lower is better.
}
\label{tab:supp_gap_l2}

\small
\setlength{\tabcolsep}{6pt}
\renewcommand{\arraystretch}{1.10}

\begin{tabular}{@{}lrrr@{}}
\toprule
\textbf{Gap (layers)}
& \textbf{Stale}
& \textbf{Absolute}
& \textbf{LTC} \\
\midrule

$5$--$8$
& 0.6914
& 0.6044
& \textbf{0.4219} \\

$9$--$12$
& 0.9363
& 1.3429
& \textbf{0.8757} \\

$13+$
& 1.0834
& 1.4216
& \textbf{1.0640} \\

\bottomrule
\end{tabular}
\end{table}

LTC reduces normalized discrepancy across all
gap ranges, although its remaining long-gap
error indicates that local transition transfer
does not exactly reconstruct the accumulated
nonlinear decoder transformations.
This is consistent with the gap-wise cosine
analysis in the main paper.


\subsection{LTC Anchor-Selection Ablation}
\label{sec:supp_anchor_ablation}

We compare random selection, spatial proximity,
semantic similarity, and their combination.
All variants use the same transition-residual
update at $K=64$, changing only the anchor
selection rule.
Definitions are provided in
\Cref{sec:supp_ablation_variants}.

\begin{table}[t]
\centering
\caption{
LTC anchor-selection ablation at $K=64$.
Semantic--spatial selection leads on three of
four benchmarks, while spatial-only selection
slightly favors RefCOCO.
All variants use the same transition-residual update.
}
\label{tab:supp_anchor_ablation}

\scriptsize
\setlength{\tabcolsep}{5pt}
\renewcommand{\arraystretch}{1.1}

\begin{adjustbox}{width=\columnwidth}
\begin{tabular}{lrrrr}
\toprule

\textbf{Anchor Selection}
& \textbf{RefCOCO} $\uparrow$
& \textbf{POPE} $\uparrow$
& \textbf{VQAv2} $\uparrow$
& \textbf{MM-Vet} $\uparrow$ \\

\midrule

Random anchors
& 19.57 & 76.23 & 70.25 & 23.58 \\

Spatial distance
& \textbf{21.39} & 76.43 & 70.32 & 22.75 \\

Semantic similarity
& 20.97 & 76.59 & 70.53 & 22.84 \\

\rowcolor{ourblue}
\textbf{Semantic + Spatial}
& 21.26
& \textbf{76.62}
& \textbf{70.56}
& \textbf{24.95} \\

\bottomrule
\end{tabular}
\end{adjustbox}
\end{table}

The results indicate that spatial and semantic
cues contribute differently across tasks.
Spatial proximity alone slightly favors RefCOCO,
whereas combining both cues performs best on
the three general multimodal benchmarks.
Thus, the combined rule provides a balanced
selection strategy rather than optimizing
every individual metric.


\subsection{Sensitivity to Routing-Layer Placement}
\label{sec:supp_routing_sensitivity}

We investigate sensitivity to the placement
of routing layers by comparing the default
schedule with three alternatives.
All configurations use the same 64-token budget
and LTC hyperparameters, differing only
in routing-layer positions.

\begin{table}[t]
\centering
\caption{
Routing-layer placement sensitivity at $K=64$.
The default schedule achieves the highest
average RefCOCO accuracy, while alternative
placements can favor POPE or VQAv2.
RefCOCO averages val, TestA, and TestB.
}
\label{tab:supp_routing_sensitivity}

\scriptsize
\setlength{\tabcolsep}{4.0pt}
\renewcommand{\arraystretch}{1.08}

\begin{adjustbox}{width=\columnwidth}
\begin{tabular}{lcrrr}
\toprule

\textbf{Routing Schedule}
& \textbf{Routing Layers}
& \textbf{RefCOCO} $\uparrow$
& \textbf{POPE} $\uparrow$
& \textbf{VQAv2} $\uparrow$ \\
\midrule

Early-concentrated
& $\{2,5,11,19\}$
& 18.26
& \textbf{77.54}
& 70.66 \\

Uniform
& $\{2,10,18,26\}$
& 18.45
& 77.01
& \textbf{71.38} \\

\rowcolor{ourblue}
\textbf{Default}
& $\mathbf{\{2,7,15,23\}}$
& \textbf{21.26}
& 76.62
& 70.56 \\

Later-shifted
& $\{3,9,17,25\}$
& 17.08
& 76.24
& 70.83 \\

\bottomrule
\end{tabular}
\end{adjustbox}
\end{table}

\Cref{tab:supp_routing_sensitivity} indicates
that grounding is more sensitive to routing-layer
placement than general multimodal performance.
In particular, schedules that improve individual
multimodal metrics do not necessarily improve
fine-grained grounding.
We use $\{2,7,15,23\}$ throughout the main
experiments as the default routing configuration.


\subsection{Additional Backbone}
\label{sec:supp_additional_backbone}

We evaluate CatchUp on Qwen2.5-VL-7B-Instruct
~\citep{bai2025qwen25vl} to assess transfer
beyond the LLaVA-1.5 backbone.
We use the same routing layers
$\{2,7,15,23\}$ and LTC hyperparameters,
comparing efficient methods at average
visual-token budgets of 64, 128, and 192.

\begin{table}[t]
\centering
\caption{
Results on Qwen2.5-VL-7B-Instruct.
CatchUp leads RefCOCO at all three average
visual-token budgets and achieves the best
or tied-best result in seven of nine
budget--benchmark comparisons.
RefCOCO averages val, TestA, and TestB.
}
\label{tab:supp_additional_backbone}

\scriptsize
\setlength{\tabcolsep}{6.0pt}
\renewcommand{\arraystretch}{1.08}

\begin{adjustbox}{width=\columnwidth}
\begin{tabular}{lrrr}
\toprule

\textbf{Method}
& \textbf{RefCOCO} $\uparrow$
& \textbf{POPE} $\uparrow$
& \textbf{VQAv2} $\uparrow$ \\
\midrule

Qwen2.5-VL-7B-Instruct
& 85.91 & 87.30 & 78.90 \\

\midrule
\multicolumn{4}{c}{
\textit{Avg. Visual Tokens = 192}
} \\
\midrule

FastV {\scriptsize (ECCV'24)}
& 46.25 & 84.63 & 71.30 \\

PyramidDrop {\scriptsize (CVPR'25)}
& 55.86 & 84.69 & 74.70 \\

ReRoute {\scriptsize (arXiv'26)}
& 58.24 & 85.66 & \textbf{77.50} \\

\rowcolor{ourblue}
\textbf{CatchUp (Ours)}
& \textbf{62.41}
& \textbf{86.02}
& \textbf{77.50} \\

\midrule
\multicolumn{4}{c}{
\textit{Avg. Visual Tokens = 128}
} \\
\midrule

FastV {\scriptsize (ECCV'24)}
& 19.62 & 79.11 & 66.30 \\

PyramidDrop {\scriptsize (CVPR'25)}
& 26.60 & 80.16 & 67.70 \\

ReRoute {\scriptsize (arXiv'26)}
& 33.68 & \textbf{82.81} & 73.90 \\

\rowcolor{ourblue}
\textbf{CatchUp (Ours)}
& \textbf{33.71}
& 82.74
& \textbf{73.97} \\

\midrule
\multicolumn{4}{c}{
\textit{Avg. Visual Tokens = 64}
} \\
\midrule

FastV {\scriptsize (ECCV'24)}
& 1.76 & 49.27 & 38.90 \\

PyramidDrop {\scriptsize (CVPR'25)}
& 10.23 & 69.63 & 57.30 \\

ReRoute {\scriptsize (arXiv'26)}
& 17.29 & 77.33 & \textbf{65.60} \\

\rowcolor{ourblue}
\textbf{CatchUp (Ours)}
& \textbf{18.13}
& \textbf{77.97}
& 65.57 \\

\bottomrule
\end{tabular}
\end{adjustbox}
\end{table}

\Cref{tab:supp_additional_backbone} shows
that the grounding benefit of CatchUp persists
on the additional backbone.
ReRoute slightly exceeds CatchUp on POPE
at 128 tokens by 0.07 points and on VQAv2
at 64 tokens by 0.03 points.
These results support transfer of the
proposed correction mechanism to a second
LVLM architecture, without establishing
architecture-independent improvements.

\subsection{Hyperparameter Sensitivity}
\label{sec:supp_hyperparameter_sensitivity}

We examine the sensitivity of CatchUp to the number of transition anchors $M$ and the spatial weight $\lambda$ at $K=64$.
We vary one hyperparameter at a time while keeping all other settings fixed.

\begin{table}[t]
\centering
\caption{
Sensitivity to the number of anchors $M$ and
spatial weight $\lambda$ at $K=64$.
The default values are $M=8$ and $\lambda=0.15$.
RefCOCO reports average Acc@0.5 over val,
TestA, and TestB.
}
\label{tab:supp_hyperparameter_sensitivity}

\scriptsize
\setlength{\tabcolsep}{5pt}
\renewcommand{\arraystretch}{1.10}

\begin{adjustbox}{max width=\columnwidth}
\begin{tabular}{llrrr}
\toprule
\textbf{Parameter}
& \textbf{Value}
& \textbf{RefCOCO} $\uparrow$
& \textbf{POPE} $\uparrow$
& \textbf{VQAv2} $\uparrow$ \\
\midrule

\multirow{3}{*}{$M$}
& 4 & 21.20 & 76.59 & 70.55 \\
& 8 & 21.26 & 76.62 & 70.56 \\
& 12 & 21.26 & 76.61 & 70.55 \\

\midrule

\multirow{3}{*}{$\lambda$}
& 0.10 & 21.18 & 76.60 & 70.56 \\
& 0.15 & 21.26 & 76.62 & 70.56 \\
& 0.20 & 21.28 & 76.57 & 70.53 \\

\bottomrule
\end{tabular}
\end{adjustbox}
\end{table}

\Cref{tab:supp_hyperparameter_sensitivity} shows
that CatchUp is relatively insensitive to moderate
variations in $M$ and $\lambda$.
Increasing $M$ beyond 8 yields no further grounding
improvement, while $\lambda=0.20$ slightly improves
RefCOCO at the expense of POPE and VQAv2.
The default configuration ($M=8$, $\lambda=0.15$)
provides balanced performance across benchmarks,
with only minor variations across the evaluated
hyperparameter ranges.


\section{Additional Efficiency Analysis}
\label{sec:supp_efficiency}

We provide detailed runtime and memory measurements
to distinguish decoder prefill savings, end-to-end
latency, and the auxiliary cost of LTC.


\subsection{Measurement Protocol}
\label{sec:supp_efficiency_protocol}

All measurements use LLaVA-1.5-7B on a single
NVIDIA RTX A6000 GPU with batch size 1
and FP16 precision.
We evaluate three fixed image--text samples,
perform two untimed warm-up passes, and collect
five timed passes per sample, yielding
15 measurements per configuration.
Each generation produces 64 tokens.

Runtime is measured using CUDA events with
explicit GPU synchronization.
We separately measure vision encoding and
projection, decoder prefill, routing,
LTC correction, and autoregressive decoding.
Residual overhead is computed as the difference
between end-to-end latency and the measured
components.
Input preprocessing is excluded from timing.

For memory analysis, we record peak PyTorch
CUDA allocated memory during each complete
64-token generation, including prefill
and decoding.
We report the mean and standard deviation
of per-generation peak allocations.

Unlike decoder KV-cache or reference-snapshot
memory, peak CUDA allocated memory captures
the combined PyTorch-managed allocation
footprint during inference.
It excludes memory allocated outside the
PyTorch CUDA allocator and does not measure
CUDA-reserved memory.


\subsection{Runtime Breakdown}
\label{sec:supp_runtime}

\begin{table*}[t]
\centering
\caption{
Component-wise inference latency on LLaVA-1.5-7B
at $K=64$ (mean $\pm$ standard deviation, ms;
15 measurements).
CatchUp reduces decoder prefill latency by 55.3\%
and end-to-end latency by 8.0\% relative to dense
inference.
Prefill excludes separately reported routing and
LTC times; other overhead is the residual.
}
\label{tab:supp_runtime_breakdown}

\scriptsize
\setlength{\tabcolsep}{5pt}
\renewcommand{\arraystretch}{1.12}

\begin{adjustbox}{max width=\textwidth}
\begin{tabular}{lcccc}
\toprule

\textbf{Component}
& \textbf{Full}
& \textbf{ReRoute}
& \textbf{w/o LTC}
& \textbf{CatchUp} \\
\midrule

Vision encoder + projector
& $40.80 \pm 1.97$
& $39.40 \pm 1.22$
& $40.20 \pm 1.85$
& $41.10 \pm 2.35$ \\

Decoder prefill
& $287.60 \pm 3.68$
& $131.70 \pm 2.21$
& $127.80 \pm 6.18$
& $128.50 \pm 5.72$ \\

Routing
& --
& $12.60 \pm 1.36$
& $10.40 \pm 0.71$
& $10.70 \pm 0.60$ \\

LTC correction
& --
& --
& --
& $14.40 \pm 1.47$ \\

Autoregressive decoding
& $1391.90 \pm 19.98$
& $1378.90 \pm 9.66$
& $1381.70 \pm 14.90$
& $1381.40 \pm 17.84$ \\

Other overhead (residual)
& $30.79 \pm 1.39$
& $30.80 \pm 1.47$
& $29.18 \pm 2.12$
& $35.07 \pm 2.74$ \\

\midrule

\textbf{E2E latency}
& $\mathbf{1751.09 \pm 22.28}$
& $\mathbf{1593.40 \pm 10.29}$
& $\mathbf{1589.28 \pm 10.20}$
& $\mathbf{1611.17 \pm 26.98}$ \\

\bottomrule
\end{tabular}
\end{adjustbox}
\end{table*}

\paragraph{Analysis.}
\Cref{tab:supp_runtime_breakdown} separates
prefill acceleration from overall inference latency.
Although token reduction substantially accelerates
prefill, autoregressive decoding accounts for
85.7\% of CatchUp's end-to-end runtime.
Consequently, reductions in prefill computation
do not translate proportionally into end-to-end
latency improvements.

LTC correction itself requires 14.40\,ms.
Relative to routing without LTC, CatchUp
increases end-to-end latency by 21.89\,ms
(1.38\%), which includes correction execution
and changes in the remaining runtime components.
Thus, representation correction incurs
a modest additional cost without replaying
the skipped decoder transformations.


\subsection{Peak CUDA Allocated Memory}
\label{sec:supp_peak_memory}

\begin{table*}[t]
\centering
\caption{
Peak PyTorch CUDA allocated memory during complete
64-token generations on LLaVA-1.5-7B
(mean $\pm$ standard deviation, MiB;
15 measurements).
CatchUp remains below dense inference at all
routing budgets, while LTC adds
21.2--23.2\,MiB over routing without correction.
The final row gives the difference in mean
peak allocation.
}
\label{tab:supp_peak_memory}

\scriptsize
\setlength{\tabcolsep}{8pt}
\renewcommand{\arraystretch}{1.12}

\begin{adjustbox}{max width=\textwidth}
\begin{tabular}{lccc}
\toprule

\textbf{Method}
& $\boldsymbol{K=64}$
& $\boldsymbol{K=128}$
& $\boldsymbol{K=192}$ \\
\midrule

Full (576 tokens)
& $18422.7 \pm 5.6$
& $18422.7 \pm 5.6$
& $18422.7 \pm 5.6$ \\

ReRoute
& $17781.7 \pm 5.6$
& $17829.4 \pm 4.2$
& $17872.1 \pm 6.5$ \\

CatchUp w/o LTC
& $17799.3 \pm 5.5$
& $17850.6 \pm 7.0$
& $17893.8 \pm 7.6$ \\

\rowcolor{ourblue}
\textbf{CatchUp}
& $17820.5 \pm 4.1$
& $17873.8 \pm 13.7$
& $17916.3 \pm 8.2$ \\

\midrule

LTC peak-allocation $\Delta$ (MiB)
& $+21.2$
& $+23.2$
& $+22.5$ \\

\bottomrule
\end{tabular}
\end{adjustbox}
\end{table*}

\paragraph{Peak allocation versus KV-cache memory.}
\Cref{tab:supp_peak_memory} shows that CatchUp
reduces peak allocated memory relative to dense
inference across all evaluated budgets.
At $K=64$, the reduction is 602.2\,MiB (3.3\%),
with corresponding savings of 548.9\,MiB
and 506.4\,MiB at $K=128$ and $K=192$.

These reductions are substantially smaller
than the decrease in active visual tokens
because model parameters and other runtime
allocations remain resident during inference.
Decoder KV-cache savings therefore cannot
be interpreted as equivalent reductions
in total PyTorch CUDA allocation.

\paragraph{LTC memory overhead.}
Enabling LTC increases mean peak allocation
by 21.2, 23.2, and 22.5\,MiB at
$K=64$, $K=128$, and $K=192$, respectively,
relative to CatchUp without LTC.

Reference-snapshot memory is a separate
auxiliary-storage metric.
At $K=64$, the measured peak reference-snapshot
storage is 22.50\,MB with LTC, compared with
0\,MB without LTC.
This snapshot-storage measurement is distinct
from the 21.2\,MiB increase in peak PyTorch
CUDA allocation; the two metrics have different
definitions and should not be equated.

CatchUp also has higher peak allocation than
ReRoute, by 38.8\,MiB at $K=64$.
Its efficiency advantage should therefore be
understood as an accuracy--resource trade-off
rather than a claim of minimum peak memory.
Nevertheless, its peak allocation remains
below dense inference at all evaluated budgets.

Together, the runtime and memory measurements
show that reversible routing reduces prefill
computation and memory requirements, while
LTC introduces a limited additional cost
to improve the representations of
re-admitted tokens.

\end{document}